\documentclass[prodmode,acmtecs]{acmsmall} 

\usepackage{graphicx}
\usepackage{epstopdf}
\usepackage{float}
\usepackage{latexsym}
\usepackage{multicol}
\usepackage{algorithmic}
\usepackage{pbox}
\usepackage{multirow}
\usepackage{hhline}
\usepackage{color}
\usepackage[table]{xcolor}
\definecolor{fluorescentpink}{rgb}{1.0, 0.08, 0.58}
\usepackage{latexsym}
\newcommand{\nop}[1]{}

\newcommand{\QED}{\mbox{\rule[0pt]{1.0ex}{1.0ex}}}
\def\boxend{\hspace*{\fill} $\QED$}

\usepackage[misc,geometry]{ifsym}

\usepackage[ruled]{algorithm2e}
\renewcommand{\algorithmcfname}{ALGORITHM}
\SetAlFnt{\small}
\SetAlCapFnt{\small}
\SetAlCapNameFnt{\small}
\SetAlCapHSkip{0pt}
\IncMargin{-\parindent}

\begin{document}
\begin{sloppy}

\begin{abstract}

Feature selection is important in many big data applications. Two critical challenges closely associate with big data. Firstly, in many big data applications, the dimensionality is extremely high, in millions, and keeps growing.  Secondly, big data applications call for highly scalable feature selection algorithms in an online manner such that each feature can be processed in a sequential scan. We present SAOLA, a \underline{S}calable and \underline{A}ccurate \underline{O}n\underline{L}ine \underline{A}pproach for feature selection in this paper. With a theoretical analysis on bounds of the pairwise correlations between features, SAOLA employs novel pairwise comparison techniques and maintain a parsimonious model over time in an online manner. Furthermore, to deal with upcoming features that arrive by groups, we extend the SAOLA algorithm, and then propose a new group-SAOLA algorithm for online group feature selection. The group-SAOLA algorithm can online maintain a set of feature groups that is sparse at the levels of both groups and individual features simultaneously. An empirical study using a series of benchmark real data sets shows that our two algorithms, SAOLA and group-SAOLA, are scalable on data sets of extremely high dimensionality, and have superior performance over the state-of-the-art feature selection methods.

\end{abstract}


\title{Scalable and Accurate Online Feature Selection for Big Data}
\author{KUI YU
\affil{Simon Fraser University}
XINDONG WU
\affil{Hefei University of Technology and University of Vermont}
WEI DING
\affil{University of Massachusetts Boston}
JIAN PEI
\affil{Simon Fraser University}
}



\keywords{Online feature selection, Extremely high dimensionality, Group features, Big data}


\begin{bottomstuff}

Author's addresses: K. Yu, School of Computing Science, Simon Fraser University, Burnaby, BC, Canada, kuiy@sfu.ca; X. Wu, Department of Computer Science, Hefei University of Technology, Hefei, 230009, China {and} Department of Computer Science, University of Vermont, Burlington, 05405, USA, email: xwu@cs.uvm.edu; W. Ding, Department of Computer Science, University of Massachusetts Boston, Boston, MA, USA, ding@cs.umb.edu; J. Pei, School of Computing Science,Simon Fraser University, Burnaby, BC, Canada, jpei@cs.sfu.ca.
\end{bottomstuff}

\maketitle
\section{Introduction}

In data mining and machine learning, the task of feature selection is to choose a subset of relevant features and remove irrelevant and redundant features from high-dimensional data towards maintaining a parsimonious model~\cite{guyon2003introduction,liu2005toward,jinxiaofs2015,zhangscaling2015}. In the era of big data today, many emerging applications, such as social media services, high resolution images, genomic data analysis, and document data analysis, consume data of extremely high dimensionality, in the order of millions or more~\cite{wu2014data,zhai2014emerging,chen2014hybrid,yu2015classification,yu2015tornado}. For example, the Web Spam Corpus 2011~\cite{wang2012evolutionary} collected approximately 16 million features (attributes) for web spam page detection, and the data set from KDD CUP 2010 about using educational data mining to accurately predict student performance includes more than 29 million features. The scalability of feature selection methods becomes critical to tackle millions of features~\cite{zhai2014emerging}.

Moreover, in many applications, feature selection has to be conducted
in an online manner. For example, in SINA Weibo, hot topics in behavior in Weibo
keep changing daily. When a novel hot topic appears, it may come with
a set of new keywords (a.k.a. a set of features). And then some of the new
keywords may serve as key features to identify new hot
topics. Another example is feature selection in bioinformatics, where
acquiring the full set of features for every training instance is
expensive because of the high cost in conducting wet lab
experiments~\cite{wang2013online}. When it is impossible to wait for a complete set of
features, it is practical to conduct feature selection from
the features available so far, and consume new features in an online
manner as they become available.

To search for a minimal subset of features that leads to the most accurate prediction model, two types of feature selection approaches were proposed in the literature, namely, batch methods~\cite{brown2012conditional,woznica2012model,javed2014correctness} and online methods~\cite{wu2012online,wang2013online}. A batch method requires loading the entire training data set into memory. This is obviously not scalable when handling large-scale data sets that exceed memory capability. Moreover, a batch method has to access the full feature set prior to the learning task~\cite{wu2012online,wang2013online}.

Online feature selection has two major approaches. One assumes that the number of features on training data is fixed while the number of data points changes over time, such as the OFS algorithm~\cite{hoi2012online,wang2013online} that performs feature selection upon each data instance. Different from OFS, the other online method assumes that the number of data instances is fixed while the number of features changes over time, such as the Fast-OSFS~\cite{wu2012online} and alpha-investing algorithms~\cite{zhou2006streamwise}. This approach maintains a best feature subset from the features seen so far by processing each feature upon its arrival. Wang et al.~\cite{wang2013online} further proposed the OGFS (Online Group Feature Selection) algorithm by assuming that feature groups are processed in a sequential scan. It is still an open research problem to efficiently reduce computational cost when the dimensionality is in the scale of millions or more~\cite{wu2012online,zhai2014emerging}.

In this paper, we propose online feature selection to tackle extremely high-dimensional data for big data analytics, our contributions are as follows.

\begin{itemize}

\item {We conduct a theoretical analysis to derive a low bound on pairwise correlations between features to effectively and efficiently filter out redundant features.}
		
\item {With this theoretical analysis, we develop SAOLA, a
  \underline{S}calable and \underline{A}ccurate
  \underline{O}n\underline{L}ine \underline{A}pproach for feature
  selection. The SAOLA algorithm employs novel online pairwise
  comparisons to maintain a
  parsimonious model over time. We analyze the
  upper bound of the gap of information gain between selected features
  and optimal features.}

\item {To deal with new features that arrive by groups, we extend the SAOLA algorithm, namely, the group-SAOLA algorithm. The group-SAOLA algorithm can online yield a set of feature groups that is sparse between groups as well as within each group for maximizing its predictive performance for classification.}

\item {An extensive empirical study using a series of benchmark data
  sets illustrates that our two methods, both SAOLA and group-SAOLA,
  are scalable on data sets of extremely high dimensionality, and have
  superior performance over the state-of-the-art online feature
  selection methods.}

\end{itemize}

The rest of the paper is organized as follows. Section 2 reviews related work. Section 3 proposes our SAOLA algorithm, and Section 4 presents the group-SAOLA algorithm. Section 5 reports our experimental results. Finally, Section 6 concludes the paper and our future work.

\section {Related Work}

Given a set of input features on a training data set, the problem of feature selection is to select a subset of relevant features from input features without performance degradation of prediction models. There are two types of feature selection approaches proposed in the literature, namely, the batch methods and the online methods.

Standard batch methods can be broadly classified into three categories: filter, wrapper and embedded methods. A wrapper method performs a forward or backward strategy in the space of all possible feature subsets, using a classifier of choice to evaluate each subset. Although this method has high accuracy, the exponential number of possible subsets makes the method computationally expensive in general~\cite{kohavi1997wrappers}. The embedded methods attempt to simultaneously maximize classification performance and minimize the number of features used based on a classification or regression model with specific penalties on coefficients of features~\cite{tibshirani1996regression,weston2000feature,zhou2011manifold}.

A filter method is independent of any classifiers, and applies evaluation measures such as distance, information, dependency, or consistency to select features~\cite{dash2003consistency,forman2003extensive,peng2005feature,song2012feature,liu2014efficient}. Then the filter methods build a classifier using selected features. Due to their simplicity and low computational cost, many filter methods have been proposed to solve the feature selection problem, such as the well-established mRMR (minimal-Redundancy-Maximal-Relevance) algorithm~\cite{peng2005feature} and the FCBF (Fast Correlation-Based Filter) algorithm~\cite{yu2004efficient}. Recently,  Zhao et al.~\cite{zhao2011similarity} proposed a novel framework to consolidate different criteria to handle feature redundancies. To bring almost two decades of research on heuristic filter criteria under a single theoretical interpretation,  Brown et al.~\cite{brown2012conditional} presented a unifying framework for information theoretic feature selection using an optimized loss function of the conditional likelihood of the training labels.

To deal with high dimensionality, Tan et al.~\cite{tan2010learning,tan2014towards} proposed the efficient embedded algorithm, the FGM (Feature Generating Machine) algorithm, and Zhai et al.~\cite{zhai2012discovering} further presented the GDM (Group Discovery Machine) algorithm that outperforms the FGM algorithm.

Group feature selection is also an interesting topic in batch methods, and it selects predictive feature groups rather than individual features. For instance, in image processing, each image can be represented by multiple kinds of groups, such as SIFT for shape information and Color Moment for color information. The lasso method (Least Absolute Shrinkage and Selection Operator) was proposed by Tibshirani~\cite{tibshirani1996regression} for shrinkage and variable selection, which minimizes the sum of squared errors with the $L_1$ penalty on the sum of the absolute values of the coefficients of features. Based on the lasso method, Yuan and Lin~\cite{yuan2006model} proposed a group lasso method to select grouped variables for accurate prediction in regression. Later, the sparse group lasso criterion as an extension of the group lasso method, was proposed by Friedman et al.~\cite{friedman2010note}, which enables to encourage sparsity at the levels of both features and groups simultaneously.

It is difficult for the standard batch method to operate on high dimensional data analytics that calls for dynamic feature selection, because the method has to access all features before feature selection can start.

Online feature selection has two research lines. One assumes that the number of features on training data is fixed while the number of data points changes over time~\cite{hoi2012online}. Recently, Wang et al.~\cite{wang2013online} proposed an online feature selection method, OFS, which assumes data instances are sequentially added.

Different from OFS, the other online approach assumes that the number of data instances is fixed while the number of features changes over time.  Perkins and Theiler~\cite{perkins2003online} firstly proposed the Grafting algorithm based on a stagewise gradient descent approach. Grafting treats the selection of suitable features as an integral part of learning a predictor in a regularized learning framework, and operates in an incremental iterative fashion, gradually building up a feature set while training a predictor model using gradient descent. Zhou et al.~\cite{zhou2006streamwise} presented Alpha-investing which sequentially considers new features as additions to a predictive model by modeling the candidate feature set as a dynamically generated stream. However, Alpha-investing requires the prior information of the original feature set and never evaluates the redundancy among the selected features.

To tackle the drawbacks, Wu et al.~\cite{wu2010online,wu2012online} presented the OSFS (Online Streaming Feature Selection) algorithm and its faster version, the Fast-OSFS algorithm. To handle online feature selection with grouped features, Wang et al.~\cite{wang2013online} proposed the OGFS (Online Group Feature Selection) algorithm. However, the computational cost inherent in those three algorithms may still be very expensive or prohibitive when the dimensionality is extremely high in the scale of millions or more.

The big data challenges on efficient online processing and scalability motivate us to develop a scalable and online processing method to deal with data with extremely high dimensionality.

\section{The SAOLA Algorithm for Online Feature Selection}

\subsection{Problem Definition}

In general, a training data set $D$ is defined by $D=\{(d_i,c_i), 1\leq i \leq N\}$, where $N$ is the number of data instances, $d_i$ is a multidimensional vector that contains $numP$ features, and $c_i$ is a class label within the vector of the class attribute $C$. The feature set $F$ on $D$ is defined by $F=\{F_1,F_2,\cdots,F_{numP}\}$. The problem of feature selection on $D$ is to select a subset of relevant features from $F$ to maximize the performance of prediction models. The features in $F$ are categorized into four disjoint groups, namely, strongly relevant, redundant, non-redundant, and irrelevant features~\cite{kohavi1997wrappers,yu2004efficient}, and the goal of feature selection is to remove redundant or irrelevant features from $F$ while keeping strongly relevant or non-redundant features. The mathematical notations used in this paper are summarized in Table~\ref{tb1}.
\begin{table}
\tbl{Summary on mathematical notations}{
\begin{tabular}{|>{}l|>{}p{9.5cm}|}
 \hline
Notation	& Mathematical meanings\\\hline
$D$	& training data set\\\hline
$F$	&input feature set on $D$\\\hline
$C$	&the class attribute\\\hline
$N$	&the number of data instances\\\hline
$numP$	&the number of features\\\hline
$G$ &the set of feature groups\\\hline
$L$ & the conditional likelihood\\\hline
$\ell$ & the conditional log-likelihood\\\hline
$G_i$ &a group of features\\\hline
$G_{t_{i-1}}$ &the set of all feature groups available till time $t_{i-1}$\\\hline
$\Psi_{t_i}$ & the set of selected groups at time $t_i$\\\hline
$S, M, S',\zeta$	&feature subsets within F\\\hline
$X, Y, Z$	&a single feature ($Y\in F,\ X\in F, Z\in F$)\\\hline
$x_i,y_i,z_i,c_i$	&an assignment of values for X, Y, Z, and C\\\hline
$s$	&an assignment of a set of values of S\\\hline
$d_i$	&a numP-dimensional data instance\\\hline
$F_i$ 	&a N-dimensional feature vector\\\hline
$f_i$	&a data value of $F_i$\\\hline
$S_{(.)}^*$	 &$S_{t_i}^*$ denote the feature subset selected at time $t_i$\\\hline
$|.|$	 &$|S_{t_i}^*|$returns the size of $S_{t_i}^*$\\\hline
$P(.|.)$ &$P(C|S)$ denotes the posterior probability of C conditioned on S\\\hline
$\delta$	&relevance threshold\\\hline
$\iota$	&the limitation of the maximum subset size\\\hline
$\alpha$	&significant level for Fisher's Z-test\\\hline
$\rho$	&p-value\\
\hline
\end{tabular}}
\label{tb1}
\end{table}

\begin{definition} [Irrelevant Feature]~\cite{kohavi1997wrappers} $F_i$ is an irrelevant feature to $C$, if and only if $\forall S\subseteq F\setminus\{F_i\}$ and $\forall f_i, \forall c_i, \forall s$ for which $P(S=s,F_i=f_i)>0$ and
$P(C=c_i|S=s,F_i=f_i)=P(C=c_i|S=s)$.
\boxend
\end{definition}

\begin{definition}[Markov Blanket~\cite{koller1995toward}] A Markov blanket of feature $F_i$, denoted as $M\subseteq F\setminus\{F_i\}$ makes all other features independent of $F_i$ given $M$, that is, $\forall Y\in F\setminus(M\cup\{F_i\})\ s.t.\ P(F_i|M,Y)=P(F_i|M)$.
\boxend
\end{definition}

By Definition 3.2, a redundant feature is defined by~\cite{yu2004efficient} as follows.

\begin{definition}[Redundant Feature] A feature $F_i\in F$ is a redundant feature and hence should be removed from $F$, if it has a Markov blanket within $F$.
\boxend
\end{definition}

We also denote $D$ by $D=\{F_i,C\}, 1\leq i \leq numP$, which is a sequence of features that is presented in a sequential order, where $F_i=\{f_1,f_2,...,f_N\}^T$ denotes the $i^{th}$ feature containing $N$ data instances, and $C$ denotes the vector of the class attribute.

If $D$ is processed in a sequential scan, that is, one dimension at a time, we can process high-dimensional data not only with limited memory, but also without requiring its complete set of features available. The challenge is that, as we process one dimension at a time, at any time $t_i$, how to online maintain a minimum size of feature subset S${_{t_i}^{\star}}$ of maximizing its predictive performance for classification. Assuming $S\subseteq F$ is the feature set containing all features available till time $t_{i-1}$,  $S{_{t_{i-1}}^{\star}}$ represents the selected feature set at $t_{i-1}$, and $F_i$ is a new coming feature at time $t_i$, our problem can be formulated as follows:

\begin{equation}\label{eq_1}
 S_{t_i}^{\star}  =  \arg\min_{S'}\{|S^{\prime}|:S^{\prime}=\mathop{\arg\max}_{\zeta\subseteq \{S{_{t_{i-1}}^{\star}}\cup F_i\}}P(C|\zeta)\}.
\end{equation}

We can further decompose it into the following key steps:

\begin{itemize}
  \item {Determine the relevance of $F_i$ to C.  Firstly, we determine whether Eq.(\ref{eq_2}) holds or not.
  \begin{equation}\label{eq_2}
 P(C|F_i)=P(C).
 \end{equation}

If so, $F_i$ is cannot add any additional discriminative power with respect to $C$, thus $F_i$ should be discarded. Hence, Eq.(\ref{eq_2}) helps us identify a new feature that either is completely useless by itself with respect to $C$, or needs to be combined with other features. If not, we further evaluate whether $F_i$ carries additional predictive information to $C$ given $S{_{t_{i-1}}^{
\star}}$, that is, whether Eq.(\ref{eq_3}) holds. If Eq.(\ref{eq_3}) holds, $F_i$ has a Markov blanket in $S{_{t_{i-1}}^{\star}}$, and thus $F_i$ should be discarded.
 \begin{equation}\label{eq_3}
 P(C|S{_{t_{i-1}}^{\star}},F_i)=P(C|S{_{t_{i-1}}^{\star}}).
 \end{equation}}
\item {Calculate $S_{t_i}^{\star}$ with the inclusion of $F_i$. Once $F_i$ is added to $S{_{t_{i-1}}^{\star}}$, at time $t_i$, $S_{t_i}$=\{$S{_{t_{i-1}}^{\star}}$, $F_i$\}, we then solve Eq.(\ref{eq_4}) to prune $S_{t_i}$ to satisfy Eq.(\ref{eq_1}).
\begin{equation}\label{eq_4}
    S{_{t_i}^{\star}}=\mathop{\arg\max}_{\zeta\subseteq S{_{t_i}}}P(C|\zeta).
    \end{equation}
}
\end{itemize}

Accordingly, solving Eq.(\ref{eq_1}) is decomposed to how to sequentially solve Eq.(\ref{eq_2}) to Eq.(\ref{eq_4}) at each time point.
Essentially, Eq.(\ref{eq_3}) and Eq.(\ref{eq_4}) deal with the problem of feature redundancy.

\subsection{Using Mutual Information to Solve Eq.(\ref{eq_1})}

To solve Eq.(1), we will employ mutual information to calculate correlations between features. Given two features $Y$ and $Z$, the mutual information between $Y$ and $Z$ is defined as follows.
\begin{equation}\label{eq_5}
I(Y;Z)=H(Y)-H(Y|Z).
\end{equation}
The entropy of feature $Y$ is defined as
\begin{equation}\label{eq_6}
H(Y)=-\Sigma_{y_i\in Y}P(y_i)\log_{2}P(y_i).
\end{equation}
And the entropy of $Y$ after observing values of another feature $Z$ is defined as
\begin{equation}\label{eq_7}
H(Y|Z)=-\Sigma_{z_j\in Z}P(z_j)\Sigma_{y_i\in Y}P(y_i|z_i)\log_{2}P(y_i|z_i),
\end{equation}
where $P(y_i)$ is the prior probability of value $y_i$ of feature $Y$, and $P(y_i|z_i)$ is the posterior probability of $y_i$ given the value $z_i$ of feature $Z$.
According to Eq.(\ref{eq_6}) and Eq.(\ref{eq_7}), the joint entropy $H(X,Y)$ between features $X$ and $Y$ is defined as follows.

\begin{equation}\label{eq_8}
\begin{array}{rcl}
H(X, Y)&=&-\Sigma_{x_i\in X}\Sigma_{y_i\in Y}P(x_i, y_i)\log_{2}P(x_i, y_i)\\
         &=&-\Sigma_{x_i\in X}\Sigma_{y_i\in Y}P(x_i, y_i)\log_{2}{P(x_i)P(y_i|x_i)}\\
         &=&-\Sigma_{x_i\in X}P(x_i)\log_{2}P(x_i)-(-\Sigma_{x_i\in X}\Sigma_{y_i\in Y}P(x_i, y_i)\log_{2}P(y_i|x_i))\\
         &=&H(X)+H(Y|X).
\end{array}
\end{equation}

From Equations~(\ref{eq_5}) to~(\ref{eq_8}), the conditional mutual information is computed by

\begin{equation}\label{eq_9}
\begin{array}{rcl}
 I(X;Y|Z)& =& H(X|Z)-H(X|YZ) \\
   &=&H(X,Z)+H(Y,Z)-H(X,Y, Z)-H(Z).
\end{array}
\end{equation}

Why can we use mutual information to solve Eq.(\ref{eq_1})? Based on
the work of~\cite{brown2012conditional}, Eq.(\ref{eq_1})
is to identify a minimal subset of features to maximize the
conditional likelihood of the class attribute $C$. Let
$S=\{S_{\theta}\cup S_{\bar{\theta}}\}$ represent the feature set
containing all features available at time $t_i$ where $S_{\theta}$
indicates the set of selected features and $S_{\bar{\theta}}$ denotes
the unselected features. Assuming $p(C|S_{\theta})$ denotes the true
class distribution while $q(C|S_{\theta})$ represents the predicted
class distribution given $S_{\theta}$, then Eq.(\ref{eq_1}) can be
reformulated as
$L(C|S_{\theta},D)=\prod_{k=1}^{N}{q(c^k|S_{\theta}^k)}$, where
$L(C|S_{\theta},D)$ denotes the conditional likelihood of the class
attribute $C$ given $S_{\theta}$ and $D$. The conditional
log-likelihood of $L(C|S_{\theta},D)$ is calculated as follows.
\begin{equation}\label{eq_10}
\ell(C|S_{\theta},D)=\frac{1}{N}\sum_{k=1}^{N}{\log{q(c^k|S_{\theta}^k)}}
\end{equation}

By the work of~\cite{brown2012conditional}, Eq.(\ref{eq_10}) can be
re-written as follows.
\begin{equation}\label{eq_11}
\ell(C|S_{\theta},D)=\frac{1}{N}\sum_{k=1}^{N}{\log{\frac{q(c^k|S_{\theta}^k)}{p(c^k|S_{\theta}^k)}}}+\frac{1}{N}\sum_{k=1}^{N}{\log{\frac{p(c^k|S_{\theta}^k)}{p(c^i|S^k)}}}+\frac{1}{N}\sum_{k=1}^{N}{\log{p(c^k|S^k)}}
\end{equation}

To negate Eq.(\ref{eq_11}) and use $E_{xy}$ to represent statistical expectation, the following equation holds\footnote{Please refer to Section 3.1 of~\cite{brown2012conditional} for the details on how to get Eq.(\ref{eq_11}) and Eq.(\ref{eq_12}).}.
\begin{equation}\label{eq_12}
-\ell(C|S_{\theta},D)=E_{xy}\bigg\{{\log{\frac{p(c^k|S_{\theta}^k)}{q(c^k|S_{\theta}^k)}}}\bigg\}+E_{xy}\bigg\{{\log{\frac{p(c^k|S^k)}{p(c^k|S_{\theta}^k)}}}\bigg\}-E_{xy}\bigg\{{\log{p(c^k|S^k)}}\bigg\}
\end{equation}

In Eq.(\ref{eq_12}), the first term is a likelihood ratio between the true and predicted class distributions given $S_{\theta}$, averaged over the input data space. The second term equals to $I(C;S_{\bar{\theta}}|S_{\theta})$, that is, the conditional mutual information between $C$ and $S_{\bar{\theta}}$, given $S_{\theta}$~\cite{brown2012conditional}. The final term is $H(C|S)$, the conditional entropy of $C$ given all features, and is an irreducible constant.

\begin{definition}[Kullback Leibler distance~\cite{kullback1951information}] The Kullback Leibler distance between two probability distributions $P(X)$ and $Q(X)$ is defined as $KL(P(X)||Q(X))=\Sigma_{x_i\in X}P(x_i)\log{\frac{P(x_i)}{Q(x_i)}}=E_{x}\log{\{\frac{P(X)}{Q(X)}\}}.$\boxend
\end{definition}

Then Eq.(\ref{eq_12}) can be re-written as follows.
\begin{equation}\label{eq_13}
\lim_{N\to\infty}{-\ell(C|S_{\theta},D)}=KL(p(C|S_{\theta})||q(C|S_{\theta}))+I(C;S_{\bar{\theta}}|S_{\theta})+H(C|S)
\end{equation}

We estimate the distribution $q(c^k|S_{\theta}^k)$ using discrete
data. The probability of a value $c^k$ of X, $p(C = x^k)$ is estimated
by maximum likelihood, the frequency of occurrences of $\{C = c^k\}$
divided by the total number of data instances $N$. Since the Strong
Law of Large Numbers assures that the sample estimate using $q$
converges almost surely to the expected value (the true distribution
$p$), in Eq.(\ref{eq_13}), $KL(p(C|S_{\theta})||q(C|S_{\theta}))$ will
approach zero with a large $N$~\cite{shlens2014notes}.

Since $I(C;S)=I(C;S_{\theta})+I(C;S_{\bar{\theta}}|S_{\theta})$ holds, minimizing $I(C;S_{\bar{\theta}}|S_{\theta})$ is equivalent to maximizing $I(C;S_{\theta})$. Accordingly, by Eq.(\ref{eq_13}), the relationship between the optima of the conditional likelihood and that of the conditional mutual information is achieved as follows.
\begin{equation}\label{eq_14}
 \arg\max_{S_{\theta}}{L(C|S_{\theta},D)}=\arg\min_{S_{\theta}}{I(C;S_{\bar{\theta}}|S_{\theta})}
\end{equation}

Eq.(\ref{eq_14}) concludes that if $I(C;S_{\bar{\theta}}|S_{\theta})$
is minimal, then $L(C|S_{\theta},D)$ is maximal. Therefore,
using mutual information as a correlation measure between
features, we propose a series of heuristic solutions to
Eq.(\ref{eq_2}), Eq.(\ref{eq_3}), and Eq.(\ref{eq_4}) in the next
section.

\subsection{The Solutions to Equations~(\ref{eq_2}) to~(\ref{eq_4})}

We can apply Definitions 3.2 and 3.3 to solve Eq.(\ref{eq_3}) and Eq.(\ref{eq_4}). However, it is computationally expensive to use Definitions 3.2 and 3.3 when the number of features within $S{_{t_{i-1}}^{\star}}$ is large. Due to evaluating whether $F_i$ is redundant with respect to $S{_{t_{i-1}}^{\star}}$ using the standard Markov blanket filtering criterion (Definitions 3.2 and 3.3), it is necessary to check all the subsets of $S{_{t_{i-1}}^{\star}}$ (the total number of subsets is $2^{|S{_{t_{i-1}}^{\star}}|}$) to determine which subset subsumes the predictive information that $F_i$ has about $C$, i.e., the Markov blanket of $F_i$. If such a subset is found, $F_i$ becomes redundant and is removed. When handling a larger number of features, it is computationally prohibitive to check all the subsets of $S{_{t_{i-1}}^{\star}}$.

Accordingly, methods such as greedy search are a natural to address this problem. In the work of~\cite{wu2012online}, a k-greedy search strategy is adopted to evaluate redundant features. It checks all subsets of size less than or equal to $\iota\ (1\leq \iota \leq |S{_{t_{i-1}}^{\star}}|)$, where $\iota$ is a user-defined parameter. However, when the size of $S_{t_{i-1}}^{\star}$ is large, it is still computationally prohibitive to evaluate the subsets of size up to $\iota$. Moreover, selecting a proper value of $\iota$ is difficult. Therefore, those challenges motivate us to develop a scalable and online processing method to solve Eq.(\ref{eq_3}) and Eq.(\ref{eq_4}) for big data analytics.

In this section, to cope with computational complexity, we propose a
series of heuristic solutions for Equations~(\ref{eq_2})
to~(\ref{eq_4}) using pairwise comparisons to calculate online the
correlations between features, instead of computing the correlations
between features conditioned on all feature subsets.

\subsubsection {Solving Eq.(\ref{eq_2})}

Assuming $S{_{t_{i-1}}^{\star}}$ is the selected feature subset at time $t_{i-1}$, and at time $t_i$, a new feature $F_i$ comes, to solve Eq.(\ref{eq_2}), given a relevance threshold $\delta$ (we will provide the detailed discussion of the parameter $\delta$ in Section 3.4.2), if $I(F_i; C)>\delta$ ($0\leq \delta<1$), $F_i$ is said to be a relevant feature to $C$; otherwise, $F_i$ is discarded as an irrelevant feature and will never be considered again.

\subsubsection{Solving Eq.(\ref{eq_3})}

If $F_i$ is a relevant feature, at time $t_i$, how can we determine whether $F_i$ should be kept given $S{_{t_{i-1}}^{\star}}$, that is, whether $I(C;F_i|S{_{t_{i-1}}^{\star}})=0$? If $\exists Y\in S{_{t_{i-1}}^{\star}}$ such that $I(F_i;C|Y)=0$, it testifies that adding $F_i$ alone to $S{_{t_{i-1}}^{\star}}$ does not increase the predictive capability of $S{_{t_{i-1}}^{\star}}$. With this observation, we solve Eq.(\ref{eq_3}) with the following lemma.

\textbf{Lemma 1} $I(X;Y|Z)\geq 0$. \boxend

\textbf{Lemma 2} With the current feature subset $S{_{t_{i-1}}^{\star}}$ at time $t_{i-1}$ and a new feature $F_i$ at time $t_i$, if $\exists Y\in S{_{t_{i-1}}^{\star}}$ such that $I(F_i;C|Y)=0$, then $I(F_i;Y)\geq I(F_i;C)$.

\textbf{Proof.} Considering Eq.(\ref{eq_5}) and Eq.(\ref{eq_9}), the following holds.

\begin{equation}\label{eq_15}
\begin{array}{rcl}
I(F_i;C)+I(F_i;Y|C)&=&H(F_i)-H(F_i|C)+H(F_i|C)-H(F_i|YC)\\
                   &=&H(F_i)-H(F_i|YC).
\end{array}
\end{equation}
\begin{equation}\label{eq_16}
\begin{array}{rcl}
I(F_i;Y)+I(F_i;C|Y)&=&H(F_i)-H(F_i|Y)+H(F_i|Y)-H(F_i|YC)\\
                   &=&H(F_i)-H(F_i|YC).
\end{array}
\end{equation}

By Equations~(\ref{eq_15}) and~(\ref{eq_16}), the following holds.
\begin{equation}\label{eq_17}
I(F_i;C|Y)=I(F_i;C)+I(F_i;Y|C)-I(F_i;Y).
\end{equation}

With Eq.(\ref{eq_17}), if $I(F_i;C|Y)=0$ holds, we get the following,
\begin{equation}\label{eq_18}
I(F_i;Y)=I(F_i;C)+I(F_i;Y|C).
\end{equation}

Using Eq.(\ref{eq_18}) and Lemma 1, the bound of $I(F_i;Y)$ is achieved.
\begin{equation}\label{eq_19}
I(F_i;Y)\geq I(F_i;C).
\end{equation}
\boxend

Lemma 2 proposes a pairwise correlation bound between features to testify whether a new feature can increase the predictive capability of the current feature subset. Meanwhile, if $I(F_i;C|Y)=0$ holds, Lemma 3 answers what the relationship between $I(Y;C)$ and $I(F_i;C)$ is.

\textbf{Lemma 3} With the current feature subset $S{_{t_{i-1}}^{\star}}$ at time $t_{i-1}$ and a new feature $F_i$ at time $t_i$, $\exists Y\in S{_{t_{i-1}}^{\star}}$, if $I(F_i;C|Y)=0$ holds, then $I(Y;C)\geq I(F_i;C)$.

\textbf{Proof.} With Eq.(\ref{eq_9}), we get $I(Y;F_i|C)=I(F_i;Y|C)$. With Eq.(\ref{eq_18}) and the following equation,
\begin{equation}\label{eq_20}
I(Y;C|F_i)-I(Y;C)=I(Y;F_i|C)-I(F_i;Y).
\end{equation}
we get the following,
$$I(Y;C|F_i)=I(Y;C)-I(F_i;C).$$
Case 1: if $I(Y;C|F_i)=0$, then the following equation holds.
\begin{equation}\label{eq_21}
I(Y;C)=I(F_i;C).
\end{equation}
Case 2: if $I(Y;C|F_i)>0$, we get the following.
\begin{equation}\label{eq_22}
I(Y;C)>I(F_i;C).
\end{equation}

By Eq.(\ref{eq_21}) and Eq.(\ref{eq_22}), Lemma 3 is proven.\boxend

According to Lemma 3, we can see that if $I(Y;C|F_i)=0$ and $I(F_i;C|Y)=0$, then $I(Y;C)$ exactly equals to $I(F_i;C)$. $F_i$ and $Y$ can replace each other. In Lemma 3, if we only consider Case 2, by Lemma 2, with the current feature subset $S{_{t_{i-1}}^{\star}}$ at time $t_{i-1}$ and a new feature $F_i$ at time $t_i$, $\exists Y\in S{_{t_{i-1}}^{\star}}$, if $I(F_i;C|Y)=0$ holds, then the following is achieved.
\begin{equation}\label{eq_23}
I(Y;C)> I(F_i;C)\ and\ I(F_i;Y)\geq I(F_i;C).
\end{equation}

With Eq.(\ref{eq_23}), we deal with Eq.(\ref{eq_3}) as follows. With a new feature $F_i$ at time $t_i$, $\exists Y\in S{_{t_{i-1}}^{\star}}$, if Eq.(\ref{eq_23}) holds, then $F_i$ is discarded; otherwise, $F_i$ is added to $S{_{t_{i-1}}^{\star}}$.

\subsubsection{Solving Eq.(\ref{eq_4})}

Once $F_i$ is added to $S_{t_{i-1}}^*$ at time $t_i$, we will check which features within $S_{t_{i-1}}^*$ can be removed due to the new inclusion of  $F_i$. If $\exists Y\in S{_{t_{i-1}}^{\star}}$ such that $I(C;Y|F_i)=0$, then Y can be removed from $S{_{t_{i-1}}^{\star}}$.

Similar to Eq.(\ref{eq_17}) and Eq.(\ref{eq_18}), if $I(C;Y|F_i)=0$, we have $I(Y;F_i)\geq I(Y;C)$. At the same time, if $I(C;Y|F_i)=0$, similar to Eq.(\ref{eq_22}), we can get,
\begin{equation}\label{eq_24}
I(F_i;C)>I(Y;C).
\end{equation}
With the above analysis, we get the following,
\begin{equation}\label{eq_25}
I(F_i;C)>I(Y;C)\ and\ I(Y;F_i)\geq I(Y;C).
\end{equation}

Accordingly, the solution to Eq.(\ref{eq_4}) is as follows. With the feature subset $S{_{t_i}^*}$ at time $t_i$ and $F_i\in S{_{t_i}^*}$, if $\exists Y\in S_{t_i}^*$ such that Eq.(\ref{eq_25}) holds, then $Y$ can be removed from $S_{t_i}^*$.

\subsection{The SAOLA Algorithm and An Analysis}

Using Eq.(\ref{eq_23}) and Eq.(\ref{eq_25}), we propose the SAOLA algorithm in detail, as shown in Algorithm 1. The SAOLA algorithm is implemented as follows. At time $t_i$, as a new feature $F_i$ arrives, if $I(F_i,C)\leq \delta$ holds at Step 5, then $F_i$ is discarded as an irrelevant feature and SAOLA waits for a next coming feature; if not, at Step 11, SAOLA evaluates whether $F_i$ should be kept given the current feature set $S_{t_{i-1}}^*$. If $\exists Y\in S{_{t_{i-1}}^{\star}}$ such that Eq.(18) holds, we discard $F_i$ and never consider it again. Once $F_i$ is added to $S_{t_{i-1}}^*$ at time $t_i$, $S_{t_{i-1}}^*$ will be checked whether some features within $S_{t_{i-1}}^*$ can be removed due to the new inclusion of  $F_i$. At Step 16, if $\exists Y\in S{_{t_{i-1}}^{\star}}$ such that Eq.(20) holds, $Y$ is removed.

\begin{algorithm}
\caption{The SAOLA Algorithm.}
\begin{algorithmic}[1]

\STATE \textbf{Input}: $F_i$:  predictive features, $C$: the class attribute; \\$\delta$: a relevance threshold ($0\leq \delta < 1$), \\$S{_{t_{i-1}}^{\star}}$:\ the selected feature set at time $t_{i-1}$;\\
 \textbf{Output}: $S_{t_i}^*$: the selected feature set at time $t_i$;

\REPEAT
\STATE get a new feature $F_i$ at time $t_i$;
\STATE /*Solve Eq.(2)*/

\IF {$I(F_i;C)\leq \delta$}
 \STATE Discard $F_i$;
 \STATE Go to Step 21;
 \ENDIF

 \FOR {each feature $Y\in S_{t_{i-1}}^*$}
 \STATE   /*Solve Eq.(3)*/

   \IF {$I(Y;C)>I(F_i;C)\ \&\ I(F_i;Y)\geq I(F_i;C)$}
      \STATE  Discard $F_i$;
       \STATE Go to Step 21;
    \ENDIF

  \STATE  /*Solve Eq.(4)*/

   \IF {$I(F_i;C)>I(Y;C)\ \&\ I(F_i;Y)\geq I(Y;C)$}
   \STATE $S_{t_{i-1}^*}=S_{t_{i-1}^*}-Y$;
   \ENDIF
\ENDFOR
 \STATE  $S_{t_i}^*=S_{t_{i-1}^*} \cup F_i$;

\UNTIL{no features are available}
\STATE Output $S_{t_i}^*$;
\end{algorithmic}
\end{algorithm}

\subsubsection{The Approximation of SAOLA}

To reduce computational cost, the SAOLA algorithm
  conducts a set of pairwise comparisons between individual features
  instead of conditioning on a set of features, as the selection
  criterion for choosing features. This is
  essentially the idea behind the well-established batch feature
  selection algorithms, such as mRMR~\cite{peng2005feature} and
  FCBF~\cite{yu2004efficient}. Due to pairwise comparisons, our
  algorithm focuses on finding an approximate Markov blanket (the
  parents and children of the class attribute in a Bayesian
  network~\cite{aliferis2010local}) and does not attempt to discover
  positive interactions between features (there exists a positive
  interaction between $F_i$ and $F_j$ with respect to $C$ even though $F_i$
 is completely useless by itself with respect to $C$, but $F_i$ can provide significantly discriminative power jointly with
  $F_j$~\cite{jakulin2003analyzing,zhao2007searching}).  In the following, we will discuss the upper bound of the gap of information gain between an approximate Markov blanket and an optimal feature set for feature selection.

Given a data set $D$, by Definition 3.2 in Section 3.1,
  if we have the optimal feature subset $M\in S$, that is the Markov
  blanket of $C$ at time $t_i$, and $S_{\theta}\in S$ is the feature
  set selected by SAOLA, and $S_{\bar{\theta}}$ represents \{$S\setminus S_{\theta}\}$, then according to the chain rule of mutual information, we get  $I((S_{\theta},S_{\bar{\theta}});C)=I(S_{\theta};C)+I(C;S_{\bar{\theta}}|S_{\theta})$. Thus, when $S_{\theta}$ takes the values of the optimal feature subset $M$, which perfectly captures the underlying distribution $p(C|M)$, then $I(C;S_{\bar{\theta}}|S_{\theta})$ would be zero. By Eq.(\ref{eq_13}), we get the following.
\begin{equation}\label{eq_26}
\begin{array}{rcl}
-\ell(C|S_{\theta},D) &=& KL(p(C|S_{\theta})||q(C|S_{\theta}))+I(C;S_{\bar{\theta}}|S_{\theta})+H(C|S)\\
&\leq &  KL(p(C|S_{\theta})||q(C|S_{\theta}))+I(C;M)+H(C|S)\\
\end{array}
\end{equation}

 Meanwhile,
in Eq.(\ref{eq_26}), the value of $KL(p(C|S_{\theta})||q(C|S_{\theta}))$ depends on how well $q$ can approximate $p$, given the selected feature set $S_{\theta}$. By Eq.(\ref{eq_13}) in Section 3.2, $KL(p(C|S_{\theta})||q(C|S_{\theta}))$ will approach zero as $N\to\infty$. Therefore, the upper bound of the gap of information gain between the selected features and optimal features can be re-written as Eq.(\ref{eq_28}). Closer $S_{\theta}$ is to $M$, smaller the gap in Eq.(27) is.
\begin{equation}\label{eq_28}
\lim_{N\to\infty}{-\ell(C|S_{\theta},D)}\leq I(C;M)+H(C|S)
\end{equation}

Our empirical results in Section 5.4 have validated that the gap between an optimal algorithm (exact Markov blanket discovery algorithms) and SAOLA, is  small using small sample-to-ratio data sets in real-world. Furthermore, SAOLA (pairwise comparisons) is much  more scalable than exact Markov blanket discovery algorithms (conditioning on all possible feature subsets) when the number of data instances or dimensionality is large.

\subsubsection{Handling Data with Continuous Values}
Finally, for data with discrete values, we use the measure of mutual information, while for data with continuous values, we adopt the best known measure of Fisher's Z-test~\cite{pena2008learning} to calculate correlations between features. In a Gaussian distribution, $Normal(\mu,\Sigma)$, the population partial correlation $p_{(F_iY|S)}$ between feature $F_i$ and  feature $Y$ given a feature subset $S$ is calculated as follows.
\begin{equation}\label{eq_30}
p_{(F_iY|S)}=\frac{-((\sum_{F_iYS})^{-1})_{F_iY}}{((\sum_{F_iYS})^{-1})_{F_iF_i}((\sum_{F_iYS})^{-1})_{YY}}
\end{equation}

In Fisher's Z-test, under the null hypothesis of
  conditional independence between $F_i$ and $Y$ given $S$,
  $p_{(F_iY|S)}=0$.  Assuming $\alpha$ is a
  given significance level and $\rho$ is the p-value returned by
  Fisher's Z-test, under the null hypothesis of the conditional
  independence between $F_i$ and $Y$, $F_i$ and $Y$ are uncorrelated
  to each other, if $\rho >\alpha$; otherwise, $F_i$ and $Y$ are
  correlated to each other, if $\rho \leq \alpha$. Accordingly, at
  time t, a new feature $F_i$ correlated to $C$ is discarded given
  $S_{t_{i-1}}^*$, if $\exists Y \in S_{t_{i-1}}^*$
  s.t. $p_{Y,C}>p_{F_i,C}$ and $p_{Y,F_i}>p_{F_i,C}$.

\subsubsection{The Parameters of SAOLA}

In Algorithm 1, we discuss the parameters used by the SAOLA algorithm in detail as follows.

\begin{itemize}

\item Relevance threshold $\delta$. It is a user-defined parameter to determine relevance thresholds between features and the class attribute.  We calculate symmetrical uncertainty~\cite{press1996numerical} instead of $I(X,Y)$, which is defined by
$$SU(X,Y)=\frac{2I(X,Y)}{H(X)+H(Y)}.$$
The advantage of $SU(X,Y)$ over $I(X,Y)$ is that $SU(X,Y)$ normalizes the value of $I(X,Y)$ between 0 and 1 to compensate for the bias of $I(X,Y)$ toward features with more values. In general, we set $0\leq \delta<1$.

\item Correlation bounds of $I(F_i;Y)$. According to Eq.(\ref{eq_23}) and Eq.(\ref{eq_25}),
  at Steps 11 and 16 of the SAOLA algorithm, $I(F_i;C)$  and  $I(Y;C)$ $(min(I(F_i;C),I(Y;C)))$ are the correlation bounds of $I(F_i;Y)$, respectively. To further validate the correlation bounds, at Steps 11 and 16, by  setting $I(Y;C)$  and $I(F_i;C)$ to $max(I(F_i;C),I(Y;C))$ respectively, we
  can derive a variant of the SAOLA algorithm, called SAOLA-max (the
  SAOLA-max algorithm uses the same parameters as the SAOLA algorithm,
  except for the correlation bounds of $I(F_i;Y)$ in Steps 11 and 16). We will conduct an empirical study on the SAOLA and SAOLA-max algorithms in Section 5.4.1.

\item Selecting a fixed number of features. For
  different data sets, using the parameters $\alpha$ or $\delta$,
  SAOLA returns a different number of selected features. Assuming the
  number of selected features is fixed to $k$, to modify our SAOLA to
  select $k$ features, a simple way is to keep the top $k$ features in
  the current selected feature set $S_{t_i}^*$ with the highest
  correlations with the class attribute while dropping the other
  features from $S_{t_i}^*$ after Step 20 in Algorithm 1.

\end{itemize}

\subsubsection{The Time Complexity of SAOLA}

The major computation in SAOLA is the computation of the correlations between features (Steps 5 and 11 in Algorithm 1). At time $t_i$, assuming the total number of features is up to $P$ and $|S_{t_i}^*|$ is the number of the currently selected feature set, the time complexity of the algorithm is $O(P|S_{t_i}^*|)$. Accordingly, the time complexity of SAOLA is determined by the number of features within $|S_{t_i}^*|$. But the strategy of online pairwise comparisons guarantees the scalability of SAOLA, even when the size of $|S_{t_i}^*|$ is large.

Comparing to  SAOLA, Fast-OSFS employs a k-greedy search strategy to filter out redundant features by checking feature subsets for each feature in $S_{t_i}^*$. At time $t_i$, the best time complexity of Fast-OSFS is $O(|S_{t_i}^*|\iota^{|S_{t_i}^*|})$, where $\iota^{|S_{t_i}^*|}$ denotes all subsets of size less than or equal to $\iota\ (1\leq \iota \leq |S{_{t_{i-1}}^{\star}}|)$ for checking. With respect to Alpha-investing, at time $t_i$, the time complexity of Alpha-investing is $O(P|S_{t_i}^*|^2)$. Since Alpha-investing only considers adding new features but never evaluates the redundancy of selected features, the feature set $S_{t_i}^*$ always has a large size. Thus, when the size of candidate features is extremely high and the size of $|S_{t_i}^*|$ becomes large, Alpha-investing and Fast-OSFS both become computationally intensive or even prohibitive. Moreover, how to select a suitable value of $\iota$ for Fast-OSFS in advance is a hard problem, since different data sets may require different $\iota$ to search for a best feature subset.

\section{A group-SAOLA Algorithm for Online Group Feature Selection}

The SAOLA algorithm selects features only at the individual feature level. When the data possesses certain group structure, the SAOLA algorithm cannot directly deal with features with   group structures. In this section, we extend our SAOLA algorithm, and propose a novel group-SAOLA algorithm to select feature groups which are sparse at the levels of both features and groups simultaneously in an online manner.

\subsection{Problem Definition}

Suppose $G=\{G_1, G_2,\cdots, G_i, \cdots , G_{numG}\}$ represents $numG$ feature groups without overlapping, and $G_i\subset F$ denotes the $i^{th}$ feature group. We denote $D$ by $D=\{G_i, C\}, 1\leq i \leq numG\}$, which is a sequence of feature groups that is presented in a sequential order. If we process those $numG$ groups in a sequential scan, at any time $t_i$, the challenge is how to simultaneously optimize selections within each group as well as between those groups to achieve a set of groups, $\Psi_{t_i}$, containing a set of selected groups that maximizes its predictive performance for classification.

Assuming $G_{i-1}\subset G$ is the set of all feature groups available till time $t_{i-1}$ and $G_i$ is a new coming group at time $t_i$, our problem can be formulated as follows:

\begin{equation}\label{eq_31}
\begin{array}{ll}
 &\Psi_{t_i}= \mathop{\arg\max}_{G_{\zeta}\subseteq \{G_{i-1}\cup G_i\}}P(C|G_\zeta) \\
 &s.t.\\
 &\ (a)\forall F_i\in G_j, G_j\subset \Psi_{t_i}, P(C|G_j\setminus\{F_i\},F_i)\neq P(C|G_j
\setminus\{F_i\}) \\
 &\ (b) \forall G_j\subset \Psi_{t_i}, P(C|\Psi_{t_i}\setminus G_j, G_j)\neq P(C|\Psi_{t_i}\setminus G_j) .
\end{array}
\end{equation}

Eq.~(\ref{eq_31}) attempts to yield a solution at time $t_i$ that is sparse at the levels of both intra-groups (constraint (a)) and  inter-groups (constraint (b)) simultaneously for maximizing its predictive performance for classification.

\begin{definition}[Irrelevant groups]
If $\exists G_i\subset G$ s.t. $I(C; G_i)=0$, then $G_i$ is considered as an irrelevant feature group.
\boxend
\end{definition}

\begin{definition}[Group redundancy in inter-groups] If $\exists G_i\subset G\ s.t.\ I(C; G_i|G\setminus G_i)=0$, then $G_i$ is a redundant group.
\boxend
\end{definition}

\begin{definition}[Feature redundancy in intra-groups] $\forall F_i\in G_i$, if $\exists S\subset G_i\setminus\{F_i\}$ s.t. $I(C;F_i|S)=0$, then $F_i$ can be removed from $G_i$.
\boxend
\end{definition}

With the above definitions, our design to solve Eq.~(\ref{eq_31}) consists of three key steps: at time $t_i$, firstly, if $G_i$ is an irrelevant group, then we discard it; if not, secondly, we evaluate feature redundancy in $G_i$ to make it as parsimonious as possible  at the intra-group level; thirdly, we remove redundant groups from the currently selected groups. The solutions to those three steps are as follows.

\begin{itemize}
\item {The solution to remove irrelevant groups.\\
At time $t_i$, if group $G_i\subset G$ comes, if $\forall F_i\in G_i$ s.t. $I(C; F_i)=0$ (or given a relevance threshold $\delta$, $I(C; F_i)\leq \delta$ ($0\leq \delta<1$), then $G_i$ is regarded as an irrelevant feature group, and thus it can be discarded.
}
\item {The solution to determine group redundancy in inter-groups.\\
Assume $\Psi_{t_i}$ is the set of groups selected at time $t_{i-1}$ and $G_i$ is a coming group at time $t_i$.  If  $\forall F_i\in G_i,\exists F_j\in G_j, and\  G_j\subset \Psi_{t_i}\ s.t.\ I(F_j; C)> I(F_i; C)$ and $I(F_j; F_i)\geq I(F_i; C)$, then $G_i$ is a redundant group, and then can be removed.
}
\item {The solution to feature redundancy in intra-groups.\\
If $G_i$ is not a redundant group, we further prune $G_i$ to make it as parsimonious as possible using Theorem 1 in Section 3.2.2.
If $\exists F_j\in G_i\ s.t.\ \exists Y\in G_i\setminus\{F_j\}, I(Y;C)> I(F_j;C)$ and $I(F_j;Y)\geq I(F_j;C)$ holds, then $F_j$ can be removed from $G_i$.}
\end{itemize}

\subsection{The Group-SAOLA Algorithm}

\begin{algorithm}
\caption{The group-SAOLA Algorithm.}
\begin{algorithmic}[1]
\STATE \textbf{Input:} $G_i$: feature group; $C$: the class attribute;\\
   $\delta$: a relevance threshold ($0\leq \delta< 1$);\\
   $\Psi_{t_{i-1}}$: the set of selected groups at time $t_{i-1}$;\\
\textbf{Output:} $\Psi_{t_i}$: the set of selected groups at time $t_i$;
\REPEAT
\STATE A new group $G_i$ comes at time $t_i$;
\STATE /*Evaluate irrelevant groups*/
 \IF {$\forall F_i\in G_i, I(F_i;C)\leq \delta$}
   \STATE Discard $G_i$;
    \STATE Go to Step 39;
 \ENDIF
 \STATE /*Evaluate feature redundancy in $G_i$*/
 \FOR {j=1 to $|G_i|$}
    \IF {$\exists Y\in \{G_i-\{F_j\}\}, I(Y;C)>I(F_j;C)\ \&\ I(Y;F_j)\geq I(F_j;C)$}
       \STATE Remove $F_j$ from $G_i$;
       \STATE Continue;
    \ENDIF
\STATE  /*Otherwise*/
   \IF {$I(F_j;C)>I(Y;C)\ \&\ I(F_j;Y)\geq I(Y;C)$}
      \STATE Remove $Y$ from $G_i$;
   \ENDIF
 \ENDFOR
\STATE /*Evaluate group redundancy in $\{\Psi_{t_{i-1}}\cup G_i\}$*/
\FOR {j=1 to $|\Psi_{t_{i-1}}|$}
   \IF{$\exists F_k\in G_j,\ \exists F_i\in G_i,\ I(F_i;C)>I(F_k;C)\ \&\ I(F_i;F_k)\geq I(F_k;C)$}
     \STATE  Remove $F_k$ from $G_j\subset \Psi_{t_{i-1}}$;
      \ENDIF
\STATE    /*Otherwise*/
    \IF{$I(F_k;C)>I(F_i;C)\ \&\ I(F_k;F_i)\geq I(F_i;C)$}
       \STATE Remove $F_i$ from $G_i$;
    \ENDIF
   \IF{$G_j$ is empty}
    \STATE $\Psi_{t_{i-1}}=\Psi_{t_{i-1}}-G_j$;
     \ENDIF
\IF{$G_i$ is empty}
      \STATE  Break;
    \ENDIF
\ENDFOR
\IF{$G_i$ is not empty}
\STATE $\Psi_{t_i}=\Psi_{t_{i-1}} \cup G_i$;
\ENDIF
\UNTIL{no groups are available}
\STATE Output $\Psi_{t_i}$;
\label{algorithm 2}
\end {algorithmic}
\end{algorithm}

With the above analysis, we propose the Group-SAOLA algorithm in Algorithm 2. In Algorithm 2, from Steps 5 to 8, if $G_i$ is an irrelevant group, it will be discarded. If not,  Step 11 and Step 16 prune $G_i$ by removing redundant features from $G_i$. At Step 22 and Step 26, the group-SAOLA removes both redundant groups in $\{\Psi_{t_{i-1}}\cup G_i\}$ and redundant features in each group currently selected. Our Group-SAOLA algorithm can online yield a set of groups that is sparse between groups as well as within each group simultaneously for maximizing its classification performance at any time $t_i$.

\section{EXPERIMENT RESULTS}

\subsection{Experiment Setup}

We use fourteen benchmark data sets as our test beds, including ten high-dimensional data sets \cite{aliferis2010local,yu2008stable} and four extremely high-dimensional data sets, as shown in Table~\ref{tb2}. The first ten high-dimensional data sets include two biomedical data sets (\emph{hiva} and \emph{breast-cancer}), three NIPS 2003 feature selection challenge data sets (\emph{dexter}, \emph{madelon}, and \emph{dorothea}), and two public microarray data sets (\emph{lung-cancer} and \emph{leukemia}), two massive high-dimensional text categorization data sets (\emph{ohsumed} and \emph{apcj-etiology}), and the \emph{thrombin} data set that is chosen from KDD Cup 2001. The last four data sets with extremely high dimensionality are available at the Libsvm data set website\footnote{http:$//$www.csie.ntu.edu.tw/$\scriptsize{\sim}$cjlin/libsvmtools/datasets/}.

In the first ten high-dimensional data sets, we use the originally provided training and validation sets for the three NIPS 2003 challenge data sets and the \emph{hiva} data set, and for the remaining six data sets, we randomly select instances for training and testing (see Table~\ref{tb2} for the number instances for training and testing.). In the \emph{news20} data set, we use the first 9996 data instances for training and the rest for testing while in the \emph{url1} data set, we use the first day data set (\emph{url1}) for training and the second day data set (\emph{url2}) for testing.  In the \emph{kdd2010} and \emph{webspam} data sets, we randomly select 20,000 data instances for training, and 100,000 and 78,000 data instances for testing, respectively. Thus, as for the~\emph{lung-cancer},~\emph{breast-cancer},~\emph{leukemia},~\emph{ohsumed},~\emph{apcj-etiology},~\emph{thrombin},~\emph{kdd10}, and~\emph{webspam} data sets, we run each data set 10 times, and report the highest prediction accuracy and the corresponding running time and number of selected features. Our comparative study compares the SAOLA algorithm with the following algorithms:

\begin{itemize}
\item {Three state-of-the-art online feature selection methods: Fast-OSFS \cite{wu2012online}, Alpha-investing \cite{zhou2006streamwise}, and OFS \cite{wang2013online}. Fast-OSFS and Alpha-investing assume features on training data arrive one by one at a time while OFS assumes data examples come one by one;}
\item {Three batch methods: one well-established algorithm of FCBF \cite{yu2004efficient}, and two state-of-the-art algorithms, SPSF-LAR \cite{zhao2011similarity} and GDM \cite{zhai2012discovering}.}
\end{itemize}
\begin{table}[ht]
\centering
\tbl{The benchmark data sets}{
\begin{tabular}{|l|r|r|r|r|}
  \hline
  Dataset & Number of features & Number of training instances & Number of testing instances\\\hline
  madelon &500 &2,000 &600\\
  hiva &1,617 &3,845 & 384\\
  leukemia &7,129 &48 &24\\
  lung-cancer&12,533 & 121 &60\\
  ohsumed & 14,373 & 3,400 &  1,600 \\
  breast-cancer &17,816 &190 & 96\\
  dexter &20,000 &300 & 300\\
  apcj-etiology	& 28,228 & 11,000 &	4,779\\
  dorothea	& 100,000 &	800	&300\\
  thrombin &	139,351	& 2,000& 543 \\\hline
  news20 &	1,355,191 &	9,996 &	10,000\\
  url1	& 3,231,961	& 20,000	& 20,000\\
  webspam &	16,609,143 & 20,000	& 78,000\\
  kdd2010 &	29,890,095 & 20,000	& 100,000\\
  \hline
\end{tabular}}
\label{tb2}
\end{table}

The algorithms above are all implemented in MATLAB except for the GDM algorithm that is implemented in C language.
We use three classifiers, KNN and J48 provided in the Spider Toolbox\footnote{http:$//$people.kyb.tuebingen.mpg.de/spider/}, and SVM\footnote{http:$//$www.csie.ntu.edu.tw/~cjlin/libsvm/} to evaluate a selected feature subset in the experiments.
The value of $k$ for the KNN classifier is set to 1 and both SVM and KNN use the linear kernel.
All experiments were conducted on a computer with Intel(R) i7-2600, 3.4GHz CPU, and 24GB memory. In the remaining sections, the parameter $\delta$ for SAOLA is set to 0 for discrete data while the significance level $\alpha$ for SAOLA is set to 0.01 for Fisher's Z-test for continuous data (the effect of $\delta$ and $\alpha$ on SAOLA was given in Section 5.4.3.). The significance level is set to 0.01 for Fast-OSFS, and for Alpha-investing, the parameters are set to the values used in~\cite{zhou2006streamwise}.

We evaluate SAOLA and its rivals based on prediction accuracy, error bar, AUC, sizes of selected feature subsets, and running time. In the remaining sections, to further analyze the prediction accuracies and AUC of SAOLA against its rivals, we conduct the following statistical comparisons.

\begin{itemize}
	\item Paired t-tests are conducted at a 95\% significance level and the win/tie/lose (w/t/l for short) counts are summarized.
\item To validate whether SAOLA and its rivals have no significant difference in prediction accuracy or AUC, we conduct the Friedman test at a 95\% significance level \cite{demvsar2006statistical}, under the null-hypothesis, which states that the performance of SAOLA and that of its rivals have no significant difference, and calculate the average ranks using the Friedman test (how to calculate the average ranks, please see~\cite{demvsar2006statistical}.).

\item When the null-hypothesis at the Friedman test is rejected, we proceed with the Nemenyi test~\cite{demvsar2006statistical} as a post-hoc test. With the Nemenyi test, the performance of two methods is significantly different if the corresponding average ranks differ by at least the critical difference (how to calculate the critical difference, please see~\cite{demvsar2006statistical}.).
\end{itemize}

We organize the remaining parts as follows. Section 5.2 compares SAOLA with online feature selection algorithms. Section 5.3 gives a comparison of SAOLA with batch feature selection methods, and Section 5.4 conducts an analysis of the effect of parameters on SAOLA. Section 5.5 compares Group-SAOLA with its rivals.

\subsection{Comparison of SAOLA with Three Online Algorithms}

\subsubsection{Prediction Accuracy, Running Time and the Number of Selected Features of SAOLA}
Since Fast-OSFS and Alpha-investing can only deal with the first ten high-dimensional data sets in Table~\ref{tb2} due to high computational cost, we compare them with SAOLA only on the first ten high-dimensional data sets.
Accordingly, in the following tables, the notation ``-'' denotes an algorithm fails to a data set because of excessive running time.

The OFS algorithm is a recently proposed online feature selection method. Since OFS uses a user-defined parameter $k$ to control the size of the final selected feature subset, we set $k$, i.e., the number of selected features to the top 5, 10, 15 ,..., 100 features, then selecting the feature set with the highest prediction accuracy as the reporting result.

\begin{table}[t]
\centering
\tbl{Prediction accuracy (J48)}{
\begin{tabular}{|l|l|l|l|l|}
\hline
Dataset & SAOLA & Fast-OSFS & Alpha-investing & OFS \\ \hline
dexter	&0.8133	&\textbf{0.8200}	&0.5000 	&0.5667\\
lung-cancer	&\textbf{0.9500}	&0.9000	&0.8333 &0.8667\\
hiva	&\textbf{0.9661}	&0.9635	&0.9635 &0.9635\\
breast-cancer	&0.7917	&\textbf{0.8854}	&\textbf{0.7187} &0.8333\\
leukemia	&\textbf{0.9583}	&\textbf{0.9583}	&0.6667 &\textbf{0.9583}\\
madelon	&0.6083	&0.6100	&0.6067 &\textbf{0.6367}\\
ohsumed	&0.9437	&\textbf{0.9450}	&0.9331 &0.9431\\
apcj-etiology	&\textbf{0.9872}	&0.9868	&0.9828 &\textbf{0.9872}\\
dorothea	&0.9343	&\textbf{0.9371}	&0.9343 &\textbf{0.9371}\\
thrombin	&\textbf{0.9613}	&0.9595	&\textbf{0.9613} &0.9374\\
news20	&\textbf{0.8276} &- &- 	&0.7332\\
url1	&\textbf{0.9744}	&- &-	&0.9720\\
kdd10	&\textbf{0.8723}	&- &-	&0.8577\\
webspam	&0.9611	&- &-	&\textbf{0.9689}\\
\hline
w/t/l  &-  &1/8/1   &5/5/0  &5/7/2\\
\hline
\end{tabular}}
\label{tb3}
\end{table}

\begin{table}[t]
\centering
\tbl{Prediction accuracy (KNN)}{
\begin{tabular}{|l|l|l|l|l|}
\hline
Dataset & SAOLA & Fast-OSFS & Alpha-investing &OFS \\ \hline
dexter	&0.7600	&\textbf{0.7800}	&0.5000 &0.5400\\
lung-cancer	&\textbf{0.9833}	&0.9667	&0.9167 &0.8500\\
hiva	&0.9635	&0.9635	&0.9531 &\textbf{0.9661}\\
breast-cancer	&\textbf{0.8646}	&0.8542	&0.6875 &0.6979\\
leukemia	 &\textbf{0.9167}	&0.7917	&0.6250 &0.8750\\
madelon	&0.5617	&0.5283	&0.5767 &\textbf{0.6433}\\
ohsumed	&0.9275	&0.9306	&\textbf{0.9325} &\textbf{0.9431}\\
apcj-etiology	&0.9793	&0.9702	&0.9851 &\textbf{0.9872}\\
dorothea	 &0.9200	&\textbf{0.9457}	&0.7400 &0.9086\\
thrombin	&0.9374	&0.9300	&0.9371 &\textbf{0.9411}\\
news20	&\textbf{0.7755}	&- &-	&0.6884\\
url1	&\textbf{0.9627}	&- &-	&0.9607\\
kdd10	&\textbf{0.8780}	&- &-	&0.7755\\
webspam	&\textbf{0.9532}	&- &-	&0.9516\\
\hline
w/t/l  &-   &4/4/2   &6/3/1  &7/5/2\\
\hline
\end{tabular}}
\label{tb4}
\end{table}

\begin{table}[t]
\centering
\tbl{Prediction accuracy (SVM)}{
\begin{tabular}{|l|l|l|l|l|}
\hline
Dataset & SAOLA & Fast-OSFS & Alpha-investing &OFS \\ \hline
dexter	&\textbf{0.8500}	&0.8100	&0.5000	&0.5000\\
lung-cancer	&\textbf{0.9833}	&0.9500	&0.9167	&0.7833\\
hiva		&\textbf{0.9635}	&\textbf{0.9635} &\textbf{0.9635} &\textbf{0.9635}\\
breast-cancer &0.8750 &\textbf{0.8854} 	&0.7188	&0.7812	\\
leukemia	 &\textbf{0.9583}	 &0.7500	 &0.6667	 &0.8333\\
madelon	&0.6217	&0.6227	&\textbf{0.6383}	&0.6117\\
ohsumed &0.9431 &\textbf{0.9438}	 &0.9431 &0.9431\\
apcj-etiology	&\textbf{0.9872}	&\textbf{0.9872}	&\textbf{0.9872}	&\textbf{0.9872}\\
dorothea	 &0.9286	 &\textbf{0.9371}	 &0.9086	 &0.9029\\
thrombin &0.9116	 &0.9116 &	0.9153	 &\textbf{0.9245}\\
news20	&\textbf{0.8721} &- &- &0.4993\\
url1	&0.9645 &- &- &\textbf{0.9681}\\
kdd10	&0.8727 &- &- &\textbf{0.8852}\\
webspam &\textbf{0.9123} &- &- &0.8897\\
\hline
w/t/l  &-  &3/6/1	&5/4/1  &8/4/2 \\
\hline
\end{tabular}}
\label{tb5}
\end{table}

Tables~\ref{tb3},~\ref{tb4}, and~\ref{tb5} summarize the prediction accuracies of SAOLA against Fast-OSFS, Alpha-investing, and OFS using the KNN, J48 and SVM classifiers. The win/tie/loss (w/t/l for short) counts of SAOLA against Fast-OSFS, Alpha-investing, and OFS are summarized in Tables~\ref{tb3},~\ref{tb4}, and~\ref{tb5}. The highest prediction accuracy is highlighted in bold face.
Tables~\ref{tb6} and ~\ref{tb7} give the number of selected features and running time of SAOLA, Fast-OSFS, Alpha-investing, and OFS. We have the following observations.

(1) SAOLA vs. Fast-OSFS. With the counts of w/t/l in Tables~\ref{tb3} and~\ref{tb5}, we observe that SAOLA is very competitive with Fast-OSFS. In Table~\ref{tb4}, we can see that SAOLA is superior to Fast-OSFS. Fast-OSFS selects fewer features than SAOLA on all data sets as shown in Table~\ref{tb6}. The explanation is that Fast-OSFS employs a k-greedy search strategy to filter out redundant features by checking the feature subsets in the current feature set for each feature while SAOLA only uses pairwise comparisons. But as shown in Table~\ref{tb7}, this strategy makes Fast-OSFS very expensive in computation and even prohibitive on some data sets, such as \emph{apcj-etiology} and the last four extremely high-dimensional data sets of Table~\ref{tb2}, as the size of the current feature set is large at each time point.

(2) SAOLA vs. Alpha-investing. From Tables~\ref{tb3} to~\ref{tb5}, we can see that SAOLA outperforms Alpha-investing on most data sets using the three classifiers. Alpha-investing selects many more features than SAOLA on~\emph{ohsumed},~\emph{apcj-etiology},~\emph{dorothea}, and~\emph{thrombin} since Alpha-investing only considers to add new features but never evaluates the redundancy of selected features. An exception is that Alpha-investing only selects one feature on the \emph{dexter} data set. A possible explanation is that the \emph{dexter} data set is a very sparse real-valued data set. Furthermore, Alpha-investing is less efficient than SAOLA as shown in Table~\ref{tb7}.
\begin{table}[ht]
\centering
\tbl{Number of selected features}{
\begin{tabular}{|l|l|l|l|l|}
\hline
Dataset & SAOLA & Fast-OSFS & Alpha-investing &OFS\\ \hline
dexter	&21	&9	&1 &85\\
lung-cancer	&35	&6	&7 &60\\
hiva	&12	&5	&48 &10\\
breast-cancer	&46	&7	&2  &10\\
leukemia	 &17	&5	&2 &45\\
madelon	&3	&3	&4  &65\\
ohsumed	&65	&11	&297 &10\\
apcj-etiology	&75	&67	&634 &10\\
dorothea	&63	&5	&113 &60\\
thrombin	&20	&9	&60 &40\\
news20	&212 &- &- &85\\
url1	&64 &- &- &100\\
kdd10	&180 &- &- &90\\
webspam &51 &- &- &85\\
\hline
\end{tabular}}
\label{tb6}
\end{table}

To validate whether SAOLA, Fast-OSFS, and Alpha-investing have no significant difference in prediction accuracy, with the Friedman test at 95\% significance level, under the null-hypothesis, which states that the performance of SAOLA and that of Fast-OSFS and Alpha-investing have no significant difference, with respect to J48 in Table~\ref{tb3}, the average ranks for SAOLA, Fast-OSFS, and Alpha-investing are 2.35, 2.40, and 1.25 (the higher the average rank, the better the performance), respectively. The null-hypothesis is rejected. Then we proceed with the Nemenyi test as a post-hoc test. With the Nemenyi test, the performance of two methods is significantly different if the corresponding average ranks differ by at least the critical difference. With the Nemenyi test, the critical difference is up to 1.047. Thus, with the critical difference and the average ranks calculated above, the prediction accuracy of SAOLA and that of Fast-OSFS have no significant difference, but SAOLA is significantly better than Alpha-investing.

As for the KNN classifier, the average ranks for SAOLA, Fast-OSFS, and Alpha-investing are 2.45, 1.85, and 1.705 in Table~\ref{tb4}, respectively. Meanwhile, as for SVM in Table~\ref{tb5}, the average ranks for SAOLA, Fast-OSFS, and Alpha-investing are 2.30, 2.25, and 1.45, respectively. Using KNN and SVM, the null-hypothesis cannot be rejected, and thus, the prediction accuracy of SAOLA and that of Fast-OSFS and Alpha-investing have no significant difference.

In summary, in prediction accuracy, SAOLA is very competitive with Fast-OSFS, and is superior to Alpha-investing. Furthermore, Fast-OSFS and Alpha-investing cannot deal with extremely high-dimensional data sets due to computational cost while SAOLA is accurate and scalable.

\begin{table}[ht]
\centering
\tbl{Running time (seconds)}{
\begin{tabular}{|l|l|l|l|l|}
\hline
Dataset & SAOLA & Fast-OSFS & Alpha-investing &OFS \\ \hline
dexter	&3	&4	&6  &\textbf{1}\\
lung-cancer	&6	&4	&2 &\textbf{1}\\
hiva	&\textbf{1}	&36	 &7  &\textbf{1}\\
breast-cancer	&5	&4	&3 &\textbf{1}\\
leukemia	 &2	&2	&1 &\textbf{0.1}\\
madelon	&\textbf{0.1}	&\textbf{0.1}	 &\textbf{0.1}  &\textbf{0.1}\\
ohsumed	&\textbf{6}	&343	&497 &9\\
apcj-etiology	&\textbf{22}	&$>3$ days	&9,781 &100\\
dorothea	 &58	&375	&457 &\textbf{10}\\
thrombin	&63	&18,576	&291 &\textbf{40}\\
news20	&\textbf{944}	&- &-	&1,572\\
url1	&\textbf{200}	&- &-	&1,837\\
kdd10	&\textbf{1,056}	&- &-	&28,536\\
webspam	&\textbf{1,456}	&- &-  &18,342\\
\hline
\end{tabular}}
\label{tb7}
\end{table}

(3) SAOLA vs. OFS. With Tables~\ref{tb3} to~\ref{tb5}, we evaluate whether the performance of SAOLA and that of OFS have no significant difference in prediction accuracy using the Friedman test at 95\% significance level. For the J48 and SVM classifiers, we observe the same average ranks for SAOLA and OFS, 1.64 and 1.36, respectively. Regarding SVM, the average ranks for SAOLA and OFS are 1.61 and 1.39, respectively. Accordingly, although SAOLA is better than OFS on prediction accuracy using the w/t/l counts and the average ranks, SAOLA and OFS have no significant difference in prediction accuracy.

However, from Table~\ref{tb6}, we can see that SAOLA selects fewer features than OSF on all data sets except for~\emph{hiva},~\emph{breast-cancer},~\emph{ohsumed},~\emph{apcj-etiology},~\emph{news20}, and~\emph{kdd10}. Moreover, Table~\ref{tb7} gives the running time of SAOLA and OFS. As for OFS, we record the running time of the feature subset with the highest accuracy as its running time. SAOLA is faster than OFS, except for the~\emph{dorothea} and~\emph{thrombin} data sets. The~\emph{dorothea} and~\emph{thrombin} data sets only include 800 samples and 2000 samples, respectively. When the number of data samples becomes large and the number of features of training data is increased to millions, OFS become very costly, and SAOLA is still scalable and efficient. The explanation is that the time complexity of SAOLA is determined by the number of features within the currently selected feature set, and the strategy of online pairwise comparisons makes SAOLA very scalable, even when the size of the current feature set is large. Moreover, setting a desirable size of a feature set selected by OFS in advance is a non-trivial task.

\subsubsection{AUC and Error Bar of SAOLA}

In this section, we further evaluate SAOLA and the three online algorithms using the error bar and AUC metrics.

\begin{table}[ht]
\centering
\tbl{AUC of SAOLA, Fast-OSFS, and Alpha-investing}{
\begin{tabular}{|l|l|l|l|}
\hline
Dataset & SAOLA & Fast-OSFS & Alpha-investing \\ \hline
dexter	 &\textbf{0.8088}	   &0.8033	   &0.5000\\
lung-cancer	&\textbf{0.9407}	   &0.9286	   &0.8298\\
hiva		&\textbf{0.5349}	   &0.5229	   &0.5326\\
breast-cancer	&\textbf{0.8111}	   &0.7996	   &0.5303\\
leukemia	 &\textbf{0.9404}	   &0.8811	   &0.6543\\
madelon	&0.5972	   &0.5883	   &\textbf{0.6072}\\
ohsumed	&0.5559	  &\textbf{0.6244}	   &0.5503\\
apcj-etiology	&\textbf{0.5256}	   &0.5250	   &0.4987\\
dorothea	 &0.7297	   &\textbf{0.8219}	   &0.6453\\
thrombin	&0.7849	   &0.7913	   &\textbf{0.8026}\\
\hline
average rank &\textbf{2.6} &2.1 &1.3\\
w/t/l &- &5/3/2 &6/2/2\\
\hline
\end{tabular}}
\label{tb8}
\end{table}

Table~\ref{tb8} reports the average AUC of J48, KNN
  and SVM for each algorithm. Using the w/t/l counts, we can see that
  our SAOLA is better than Fast-OSFS and Alpha-investing. To further
  validate whether SAOLA, Fast-OSFS, and Alpha-investing have no
  significant difference in AUC, with the Friedman test, the
  null-hypothesis is rejected, and the average ranks calculated for
  SAOLA, Fast-OSFS, and Alpha-investing are 2.6, 2.1, and 1.3,
  respectively.

Then we proceed with the Nemenyi test as a post-hoc test. With the Nemenyi test, the performance of two methods is significantly different if the corresponding average ranks differ by at least the critical difference. With the Nemenyi test, the critical difference is up to 1.047. Accordingly, the AUC of SAOLA and that of Fast-OSFS have no significant difference, but SAOLA is significantly better than Alpha-investing.

Figure~\ref{fig1} shows the average AUC of J48, KNN and SVM and its standard deviation for SAOLA and OFS. We can see that SAOLA outperforms to OFS on all 14 data sets.

From Table~\ref{tb8} and Figure~\ref{fig1}, we can conclude that none of SAOLA, Fast-OSFS, Alpha-investing, and OFS can effectively deal with highly class-imbalanced data sets, such as~\emph{hiva},~\emph{apcj-etiology}, and~\emph{ohsumed}.

\begin{figure}
\centering
\includegraphics[height=2in]{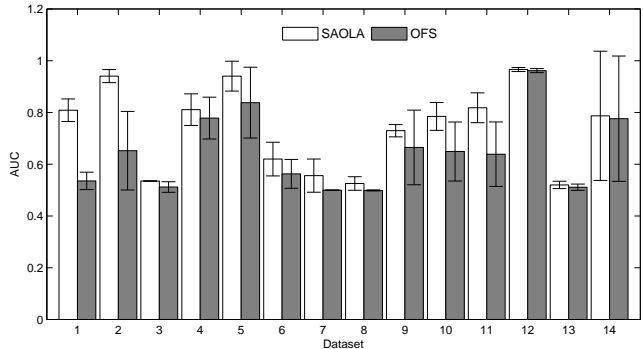}
\caption{The AUC of SAOLA and OFS (the labels of the x-axis from 1 to 14 denote the data sets: 1. dexter; 2. lung-cancer; 3. hiva; 4. breast-cancer; 5. leukemia; 6. madelon; 7. ohsumed; 8. apcj-etiology; 9. dorothea; 10. thrombin; 11. news20; 12. url1; 13. kdd10; 14. webspam)}
\label{fig1}
\end{figure}

Finally, we give the error bars of SAOLA, Fast-OSFS,
  Alpha-investing, and OFS. Since we randomly select instances from
  the~\emph{lung-cancer},~\emph{breast-cancer},~\emph{leukemia},~\emph{ohsumed},~\emph{apcj-etiology},~\emph{thrombin},~\emph{kdd10},
  and~\emph{webspam} data sets for training and testing, we only
  report the error bars of the four online algorithms on those eight data
  sets using the SVM classifier. For each data set, we randomly run
  each algorithm 10 times, and then compute the average prediction
  accuracy and the corresponding standard deviation. Figure~\ref{fig2}
  plots the average prediction accuracy and an error bar that denotes
  a distance of the standard deviation above and below this average
  prediction accuracy on each data set. Figure~\ref{fig2} shows that
  SAOLA achieves a higher average prediction accuracy and a lower
  standard deviation than Alpha-investing and OFS while being highly
  comparable with Fast-OSFS. This further confirms that SAOLA is very
  competitive with Fast-OSFS and superior to Alpha-investing and
  OFS.

\begin{figure}
\centering
\includegraphics[height=1.5in]{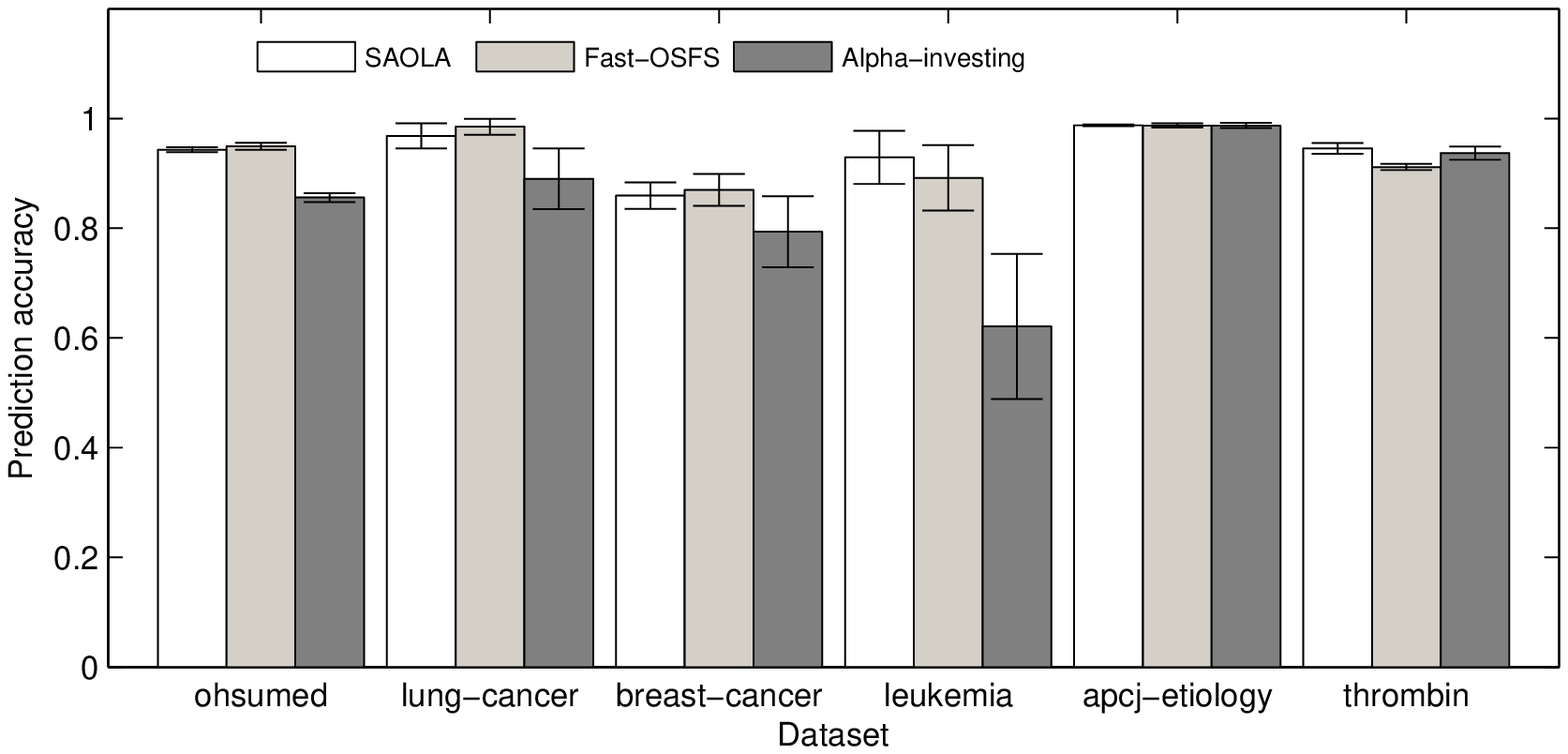}
\includegraphics[height=1.5in]{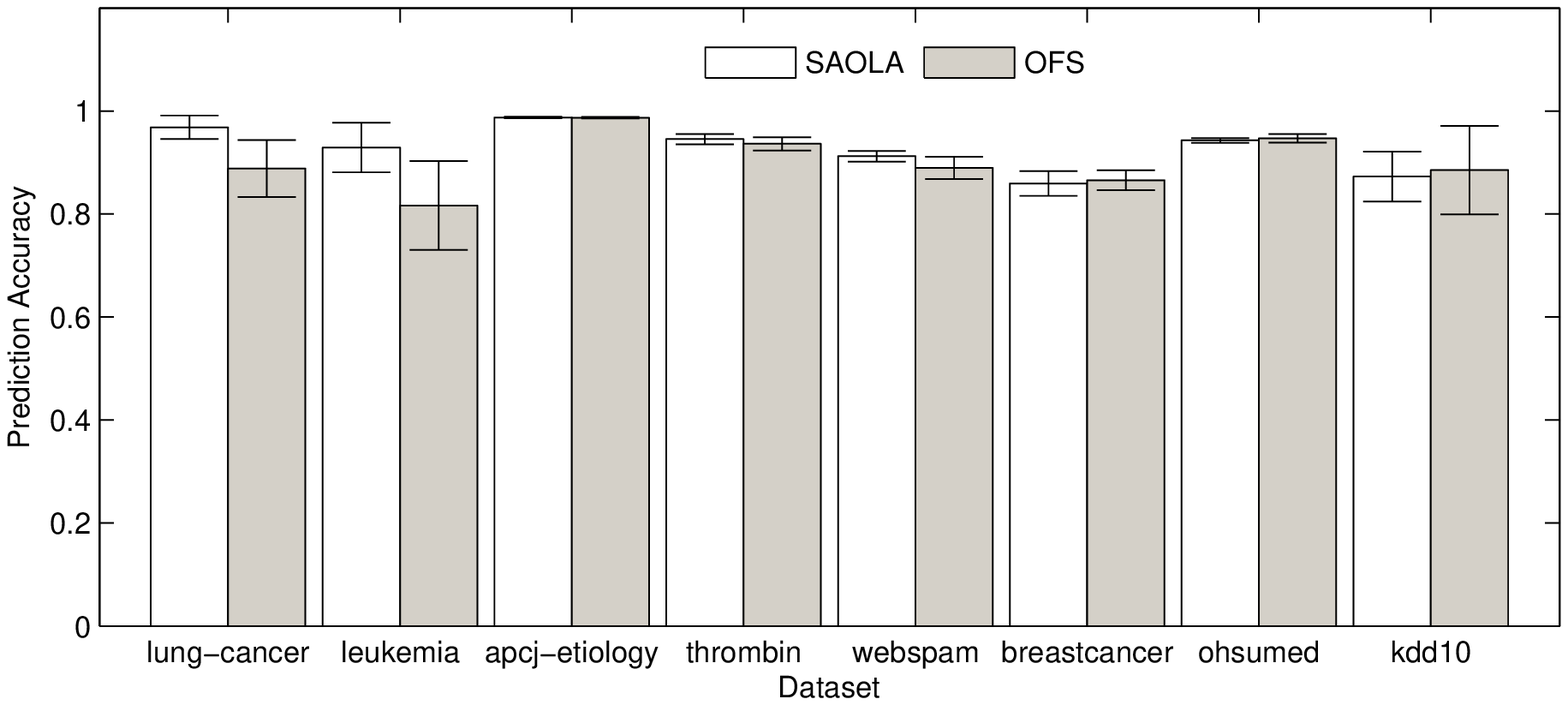}
\caption{The error bars of SAOLA, Fast-OSFS, Alpha-investing, and OFS}
\label{fig2}
\end{figure}

\subsubsection{Stability of SAOLA}

The stability of feature selection is one of the criteria to measure the performance of a feature selection algorithm by quantifying the `similarity' between two selected feature sets, and was first discussed by Kalousis et al.~\cite{kalousis2007stability}. In this section, we employ the measure proposed by Yu et al.~\cite{yu2008stable} to evaluate the stabilities of SAOLA, Fast-OSFS, Alpha-investing, and OFS. This measure constructs a weighted complete bipartite graph, where the two node sets correspond to two different feature sets, and weights assigned to the arcs are the normalized mutual information between the features at the nodes, also sometimes referred to as the symmetrical uncertainty. The Hungarian algorithm is then applied to identify the maximum weighted matching between the two node sets, and the overall similarity between two sets is the final matching cost.

To evaluate the stabilities, each data set was randomly partitioned into five folds, each fold containing 1/5 of all the samples. SAOLA and its rivals under comparison were repeatedly applied to four out of the five folds. This process was repeated 30 times to generate different subsamples for each data set. Then the average stabilities over 30 subsamples on each data set are as the results of SAOLA, Fast-OSFS, Alpha-investing, and OFS.

Figure~\ref{fig4} shows the stabilities of SAOLA, Fast-OSFS, and Alpha-investing (we do not plot the stability of Alpha-investing on the~\emph{dexter} data set, since Alpha-investing only selects one feature on the~\emph{dexter} data set.). We can see that the Alpha-investing algorithm is the most stable feature selection algorithm among the three online methods. The explanation is that the Alpha-investing algorithm only considers adding features while never removes redundant features. SAOLA and Fast-OSFS aim to select a minimum subset of features necessary for constructing a classifier of best predictive accuracy and discard features which are relevant to the class attribute but highly correlated to the selected ones. Among a set of highly correlated features, different ones may be selected under different settings of SAOLA and Fast-OSFS. Therefore, from Figure~\ref{fig4}, we can see that SAOLA is very competitive with Fast-OSFS.

Meanwhile, from Figure~\ref{fig5}, we can observe that SAOLA is more stable than OFS. Such observation illustrates that even if OFS can select large subsets of features, it is still less stable than SAOLA. The possible explanation is that OFS assumes that data examples come one by one while SAOLA assume that features on training data arrive one by one at a time.
\begin{figure}
\centering
\includegraphics[height=2in]{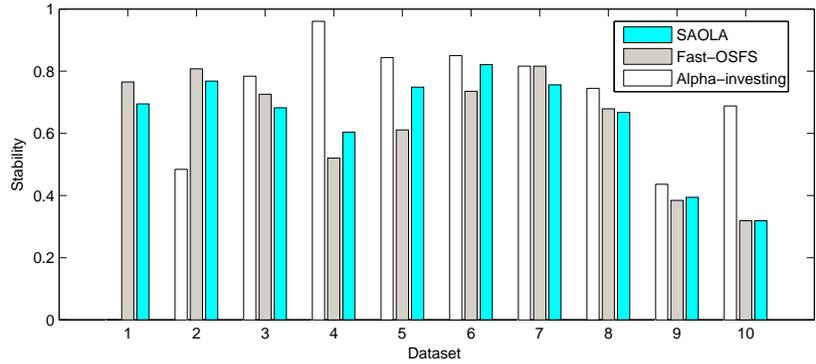}
\caption{The stabilities of SAOLA, Fast-OSFS, and Alpha-investing  (The labels of the x-axis from 1 to 10 denote the data sets: 1. dexter; 2. lung-cancer; 3. hiva; 4. breast-cancer; 5. leukemia; 6. madelon; 7. ohsumed; 8. apcj-etiology; 9. dorothea; 10. thrombin)}
\label{fig4}
\end{figure}
\begin{figure}
\centering
\includegraphics[height=2in]{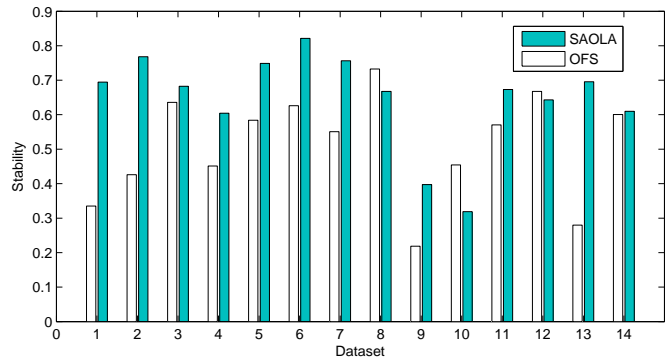}
\caption{The stabilities of SAOLA and OFS (The labels of the x-axis from 1 to 14 denote the data sets: 1. dexter; 2. lung-cancer; 3. hiva; 4. breast-cancer; 5. leukemia; 6. madelon; 7. ohsumed; 8. apcj-etiology; 9. dorothea; 10. thrombin; 11.news20; 12. url; 13. kdd10; 14. webspam)}
\label{fig5}
\end{figure}

\subsubsection{The Effect of Large Data Sets on SAOLA}

To evaluate the effect of large data sets on SAOLA, we use the data sets,~\emph{connect-4} (60,000/7,557 objects and 126 attributes) (note: 60,000/7,557 objects denote 60,000 data instances for training while 7,557 ones for testing),~\emph{ijcnn1} (190,000/1,681 objects and 22 attributes),~\emph{covtype} (571,012/10,000 objects and 54 attributes),~\emph{poker} (1,020,000/5,010 objects and 11 attributes), and~\emph{real-sim} (70,000/2,309 objects and 20,958 attributes) from the machine learning data set repository~\footnote{http:$//$mldata.org/repository/data/}.

Table~\ref{new_tb11} gives the running time of SAOLA
  and its rivals. In Table~\ref{new_tb11}, ``-'' denotes that an algorithm
  fails to the data set due to the high computational cost
  (exceeding three days). We can see that SAOLA is more scalable to
  deal with data with large numbers of data instances than Fast-OSFS,
  Alpha-investing, and OFS. Furthermore, Figure~\ref{new_fig1} gives
  the prediction accuracy of the four algorithms using the SVM
  classifier. Since Fast-OSFS fails on the ~\emph{connect-4}
  and~\emph{real-sim} data sets while Alpha-investing cannot run
  on the~\emph{real-sim} data set, Figure~\ref{new_fig1} does not plot the  prediction accuracies of Fast-OSFS and Alpha-investing on those  data sets.  SAOLA is better than Fast-OSFS and OFS and competitive with Alpha-investing on prediction accuracy.

\begin{table}[ht]
\centering
\tbl{Running time of SAOLA and its three rivals (in seconds)}{
\begin{tabular}{|l|l|l|l|l|}
\hline
Dataset & SAOLA & Fast-OSFS & Alpha-investing &OFS\\ \hline
connect-4 &2 &- &38 &310\\
ijcnn1 &0.45 &2 &0.75 &2\\
covtype &13 &15,324 &48 &8,846\\
poker   &0.26 &0.42 &0.92 &10\\
real-sim &1,223 &- &- &1,013\\
\hline
\end{tabular}}
\label{new_tb11}
\end{table}

\begin{figure}
\centering
\includegraphics[height=1.5in,width=3.5in]{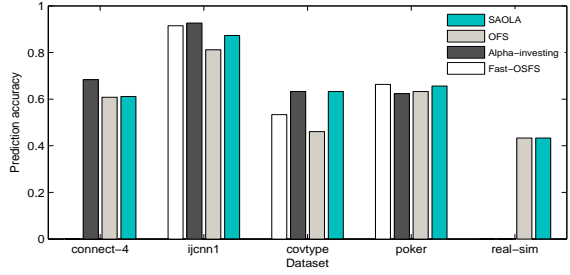}
\caption{Prediction accuracy of SAOLA, Fast-OSFS, Alpha-investing, and OFS}
\label{new_fig1}
\end{figure}

\subsection{Comparison with the Three Batch Methods}

\subsubsection{Running Time, the Number of Selected Features, and Prediction Accuracy of SAOLA}

Since FCBF and SPSF-LAR can only deal with the first ten high-dimensional data sets in Table~\ref{tb2}, in the following tables and figures, we compare FCBF and SPSF-LAR with our proposed algorithm only on those ten high-dimensional data sets in terms of size of selected feature subsets, running time, and prediction accuracy. The information threshold for FCBF is set to 0. We set the user-defined parameter k, i.e., the number of selected features to the top 5, 10, 15 ,..., 65 features for the SPSF-LAR algorithm, choose the feature subsets of the highest prediction accuracy, and record the running time and the size of this feature set as the running time and the number of selected features of SPSF-LAR, respectively.

We also select the GDM algorithm~\cite{zhai2012discovering} which is one of the most recent batch feature selection methods in dealing with very large dimensionality. The GDM algorithm is an efficient embedded feature selection method using cutting plane strategy and correlation measures as constraints to minimize the correlation among the selected features.
GDM uses a user-defined parameter to control the size of the final selected feature subset. We set the selected feature subset sizes to the top 10, 20, 30, ..., 260 features for the GDM algorithm, report the running time of the feature subset with the highest accuracy as the running time of GDM, and choose the highest prediction accuracies achieved among those selected feature subsets.

\begin{figure}[t]
\centering
\includegraphics[height=1.5in,width=3.2in]{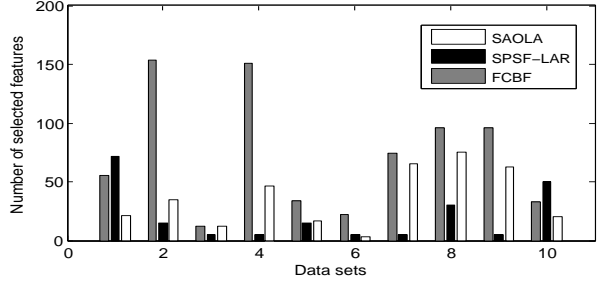}
\caption{Number of selected features (The labels of the x-axis from 1 to 10 denote the data sets: 1. dexter; 2. lung-cancer; 3. hiva; 4. breast-cancer; 5. leukemia; 6. madelon; 7. ohsumed; 8. apcj-etiology; 9. dorothea; 10. thrombin)}
\label{fig6}
\end{figure}

\begin{figure}[t]
\centering
\includegraphics[height=1.5in,width=3.2in]{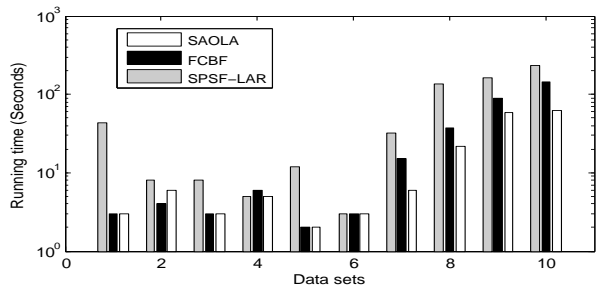}
\caption{Running time (the labels of the x-axis are the same as the labels of the x-axis in Figure~\ref{fig6}.)}
\label{fig7}
\end{figure}

From Figure~\ref{fig6}, we can conclude that FCBF
  selects the most features among SAOLA, FCBF and SPSF-LAR while SAOLA
  and SPSF-LAR are similar to each other. As shown in
  Figure~\ref{fig7}, we can observe that SAOLA is the fastest
  algorithm among SAOLA, FCBF and SPSF-LAR while SPSF-LAR is the
  slowest. The explanation is that the time complexity of the
  algorithm is $O(numP|S_{t_i}^*|)$ where $numP$ is the number of features
  and $|S_{t_i}^*|$ is the number of selected features at time $t_i$,
  while the time complexity of SPFS-LAR is $O(numPNk + Nk^3)$ where $N$
  is the number of data samples and $k$ the number of selected
  features.

The computational costs of SAOLA and FCBF are very
  competitive since both of them employ pairwise comparisons to
  calculate the correlations between features. But when the number of
  data instances or the number of features is large, SAOLA is faster
  than FCBF, such as on~\emph{ohsumed},~\emph{apcj-etiology},~\emph{dorothea}, and
  ~\emph{thrombin}. A possible reason is that SAOLA online evaluates
  both the new coming features (Step 11) and the current feature set
  (Step 16) at each time point to make the selected feature set as
  parsimonious as possible, since the size of the selected feature set
  at each time point is the key to determine the running time of
  SAOLA. With the same information threshold $\delta$, FCBF prefers
  to selecting more features than SAOLA.

Figure~\ref{fig8} shows the running time of SAOLA against GDM. Since GDM is implemented in C++, we developed a C++ version of SAOLA for the comparison with GDM, in addition to its Matlab version. In Figure~\ref{fig8}, we only give the last four data sets with extremely high dimensionality in Table~\ref{tb2}, since on the the first ten data sets, the running time of both SAOLA and GDM is no more than ten seconds. We can see that although GDM is a wrapper-like feature selection method, both GDM and SAOLA are very efficient to handle extremely high-dimensional data sets. Except for the~\emph{news20} data set, SAOLA is a little faster than GDM. On the sparse data sets, SAOLA is faster than GDM, while on the dense data sets, such as the~\emph{news20} data set, GDM is faster than SAOLA. Finally, Figure~\ref{fig9} reports the number of selected features of SAOLA comparing to GDM. Except for the~\emph{breast-cancer} data set, SAOLA selects fewer features than GDM to achieve the very competitive prediction accuracy with GDM.

\begin{figure}
\centering
\includegraphics[height=1.4in,width=3.4in]{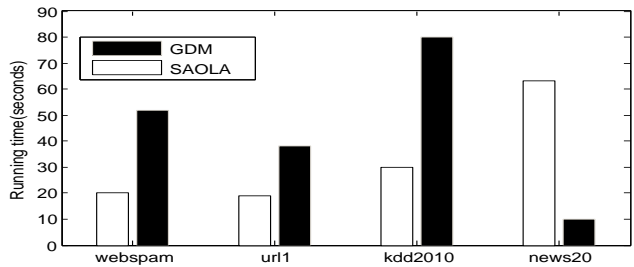}
\caption{Running time of SAOLA and GDM}
\label{fig8}
\end{figure}

\begin{figure}
\centering
\includegraphics[height=1.5in,width=3.4in]{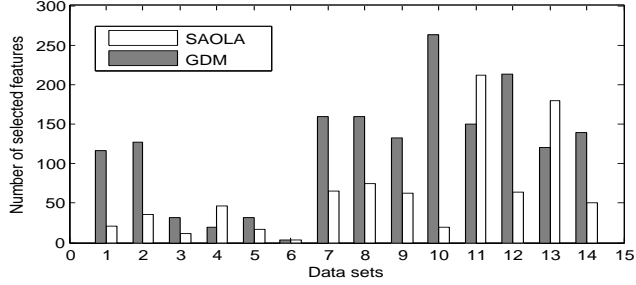}
\caption{Number of selected features (The labels of the x-axis from 1 to 14 denote the data sets: 1. dexter; 2. lung-cancer; 3. hiva; 4. breast-cancer; 5. leukemia; 6. madelon; 7. ohsumed; 8. apcj-etiology; 9. dorothea; 10. thrombin; 11. news20; 12. url1; 13. kdd10; 14. webspam)}
\label{fig9}
\end{figure}

Finally, Tables~\ref{tb9} to~\ref{tb11} report the prediction accuracies of SAOLA against FCBF, SPSF-LAR, and GDM. With the counts of w/t/l in the last rows of Tables~\ref{tb9} to~\ref{tb11}, we can see that even without requiring the entire feature set on a training data set in advance, SAOLA is still very competitive with FCBF, SPSF-LAR and GDM in prediction accuracy.

Using the Friedman test at 95\% significance level, for J48, the average ranks for SAOLA, FCBF and SPSF-LAR are 2.05, 1.90, and 2.05, respectively. For KNN, the average ranks for SAOLA, Fast-OSFS, and SPSF-LAR are 1.90, 2.25 and 1.85, respectively, while regarding SVM, the average ranks are 2.15, 1.65, and 2.20. Thus, with the Friedman test at 95\% significance level, using KNN, J48 and SVM, the null-hypothesis cannot be rejected, and thus SAOLA, FCBF and SPSF-LAR have no significant difference in prediction accuracy. Accordingly, we conclude that the performance of SAOLA is highly comparable to that of FCBF and SPSF-LAR.

\begin{table}[t]
\centering
\tbl{Prediction accuracy (J48)}{
\begin{tabular}{|l|l|l|l|l|}
\hline
Dataset & SAOLA & FCBF & SPSF-LAR &GDM\\ \hline
dexter	&0.8133	&0.8567	&0.8700 &\textbf{0.9100}\\
lung-cancer	&0.9500	&0.9500	&\textbf{0.9833} &\textbf{0.9833}\\
hiva	 &\textbf{0.9661}	 &\textbf{0.9661}	&0.9635 &\textbf{0.9661}\\
breast-cancer	&0.7917	&0.8125	&\textbf{0.8958} &0.4792\\
leukemia	 &0.9583 	&0.9583	&0.9583 &\textbf{1.0000}\\
madelon	&0.6083	&0.6067	&\textbf{0.6183} &0.5833\\
ohsumed	&0.9437	&\textbf{0.9444} &0.9431 &0.9438\\
apcj-etiology &0.9872	&0.9866	&0.9872 &\textbf{0.9879}\\
dorothea	 &0.9343 &0.9314	 &0.9029 &\textbf{0.9371}\\
thrombin	&\textbf{0.9613}	&0.9576	&0.9558 &0.7300\\
news20	&\textbf{0.8276}	&- &- &0.7354\\
url1	&0.9744	&- &- &\textbf{0.9765}\\
kdd10	&0.8723	&- &- &\textbf{0.8779}\\
webspam	&0.9611	&- &-  &\textbf{0.9617}\\
\hline
w/t/l  &-   &0/8/2 &1/5/4  &4/7/3\\
\hline
\end{tabular}}
\label{tb9}
\end{table}

\begin{table}[t]
\centering
\tbl{Prediction accuracy (KNN)}{
\begin{tabular}{|l|l|l|l|l|}
\hline
Dataset & SAOLA & FCBF & SPSF-LAR &GDM\\ \hline
dexter	&0.7600	&0.7967	&0.7233 &\textbf{0.9100} \\
lung-cancer	&\textbf{0.9833}	&0.9500	&\textbf{0.9833} &\textbf{0.9833}\\
hiva	&\textbf{0.9635}	&0.9609	&0.9635 &\textbf{0.9661}\\
breast-cancer	&\textbf{0.8646}	&0.8333	&0.8229 &0.4792\\
leukemia	 &0.9167	&\textbf{1.0000}	&\textbf{1.0000} &\textbf{1.0000}\\
madelon	&0.5617	&0.5767	&0.5633 &\textbf{0.5833}\\
ohsumed	&0.9275	&0.9300	&0.9113 &\textbf{0.9438}\\
apcj-etiology	&0.9793	&0.9826	&0.9803 &\textbf{0.9879}\\
dorothea	 &0.9200	&0.9200	&0.8857 &\textbf{0.9371}\\
thrombin	&0.9374	&0.9429	&\textbf{0.9650} &0.7300\\
news20	&\textbf{0.7755}	&- &- &0.7354\\
url1	&0.9627	&- &- &\textbf{0.9765}\\
kdd10	&\textbf{0.8780}	&- &- &0.8779\\
webspam	&0.9532	&- &-  &\textbf{0.9617}\\
\hline
w/t/l  &-   &2/5/3   &4/4/2 &3/5/6\\
\hline
\end{tabular}}
\label{tb10}
\end{table}

\begin{table}[t]
\centering
\tbl{Prediction accuracy (SVM)}{
\begin{tabular}{|l|l|l|l|l|}
\hline
Dataset & SAOLA & FCBF & SPSF-LAR &GDM\\ \hline
dexter	&0.8500	&0.5400	&0.6400 &\textbf{0.9100} \\
lung-cancer	&\textbf{0.9833}	&0.9800	&\textbf{0.9833} &\textbf{0.9833}\\
hiva	 &0.9635	&0.9635	&0.9635 &\textbf{0.9661}\\
breast-cancer &0.8750	 &0.8750	 &\textbf{0.8854}  &0.4792\\
leukemia	 &0.9583 &	0.9167	 &0.9167  &\textbf{1.0000}\\
madelon	  &0.6217	&0.5933	&\textbf{0.7900} &0.5833\\
ohsumed	&0.9431	&0.9431	&0.9431 &\textbf{0.9438}\\
apcj-etiology	&0.9872	&0.9872	&0.9872 &\textbf{0.9879}\\
dorothea	 	 &0.9286	 &0.9171	 &0.9029 &\textbf{0.9371}\\
thrombin	&\textbf{0.9116}	&\textbf{0.9116}	&0.9153 &0.7300\\
news20  &\textbf{0.8721}	&- &- &0.7354\\
url1	&0.9645	&- &- &\textbf{0.9765}\\
kdd10	&0.8727	&- &- &\textbf{0.8779}\\
webspam	&0.9123	&- &-  &\textbf{0.9617}\\
\hline
w/t/l  &-   &4/6/0   &3/5/2 &4/6/4\\
\hline
\end{tabular}}
\label{tb11}
\end{table}

With regard to the prediction accuracies of SAOLA and GDM, we can see that our algorithm is very competitive with GDM on J48, KNN, and SVM. With the Friedman test at 95\% significance level, for J48, the average ranks for SAOLA and GDM are 1.32 and 1.68, respectively. As for KNN, the average ranks for SAOLA and GDM are 1.39 and 1.61, respectively. Using KNN and J48, the null-hypothesis cannot be rejected, accordingly, the SAOLA and GDM do not have significant difference in prediction accuracy.

As for SVM, the null-hypothesis is rejected, and the average ranks for SAOLA and GDM are 1.25 and 1.75, respectively. Then we proceed with the Nemenyi test as a post-hoc test. With the Nemenyi test, the critical difference is up to 0.5238. Thus, with the critical difference and the average ranks calculated above, GDM is significantly better than SAOLA. From the results above, we can see that the GDM algorithm is inferior to SAOLA on some data sets, such as~\emph{thrombin} and~\emph{news20}, since those data sets are very sparse. However, Fisher's Z-test and information gain employed by SAOLA can deal with those spare data sets well.

In summary, our SAOLA algorithm is a scalable and accurate online approach. Without requiring a complete set of features on a training data set before feature selection starts, SAOLA is very competitive with the well-established and state-of-the-art FCBF, SPSF-LAR, and GDM methods.

\subsubsection{AUC and Error Bar of SAOLA}

In this section, we compare SAOLA with the three batch algorithms using the error bar and AUC metrics.

\begin{table}[ht]
\centering
\tbl{AUC of SAOLA, FCBF, and SPSF-LAR}{
\begin{tabular}{|l|l|l|l|}
\hline
Dataset & SAOLA &FCBF & SPSF-LAR \\ \hline
dexter	&\textbf{0.8088}	   &0.7311	   &0.7611 \\
lung-cancer	&0.9407	   &0.9475	   &\textbf{0.9639}\\
hiva		&\textbf{0.5349}	   &0.5347	   &0.5335\\
breast-cancer	&0.8111	   &0.8120	   &\textbf{0.8466}\\
leukemia	&0.9404	   &\textbf{0.9737}	&\textbf{0.9737} \\
madelon	&0.5972	   &0.5962	   &\textbf{0.7537}\\
ohsumed	&0.5559	   &\textbf{0.5741}	   &0.5613\\
apcj-etiology	&0.5256	   &\textbf{0.5473}	   &0.4987\\
dorothea	 &\textbf{0.7297}	   &0.6775	   &0.7218\\
thrombin	&0.7849	   &0.7938	   &\textbf{0.8382}\\
\hline
average rank &1.8 &1.95 &\textbf{2.15}\\
w/t/l &- &3/4/3 &2/3/5\\
\hline
\end{tabular}}
\label{tb12}
\end{table}

Table~\ref{tb12} reports the average AUC of J48, KNN
  and SVM for each algorithm. Using the w/t/l counts, we can see that
  our SAOLA is very competitive with FCBF and SPSF-LAR. To further
  validate with the Friedman test and the null-hypothesis, the average ranks calculated  for SAOLA, FCBF and SPSF-LAR are 1.8, 1.95, and 2.15,
  respectively. Accordingly, the AUC of SAOLA and that of FCBF and
  SPSF-LAR have no significant difference.

Figure~\ref{fig10} shows the average AUC of J48, KNN
  and SVM and its standard deviation for SAOLA and GDM. From
  Figure~\ref{fig10}, GDM outperforms SAOLA on most data sets,
  but the AUC of SAOLA is highly comparable to that of GDM. Moreover,
  we can see that none of these algorithms, SAOLA, FCBF, SPSF-LAR, and GDM,
  can effectively deal with highly class-imbalanced data sets.

\begin{figure}
\centering
\includegraphics[height=1.8in]{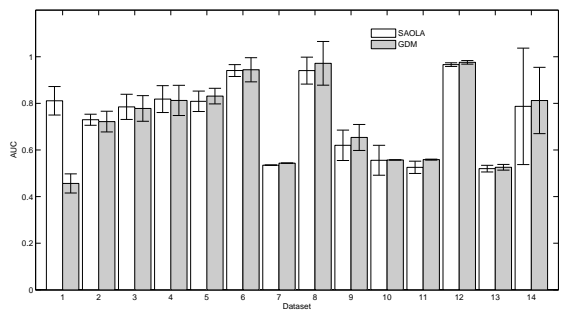}
\caption{The AUC of SAOLA and GDM  (The labels of the x-axis from 1 to 14 denote the data sets: 1. breast-cancer; 2. dorothea; 3. thrombin; 4. news20; 5. dexter; 6. lung-cancer; 7. hiva; ; 8. leukemia; 9. madelon; 10. ohsumed; 11. apcj-etiology; 12. url1; 13. kdd10; 14. webspam)}
\label{fig10}
\end{figure}

We give the error bars of SAOLA, FCBF, SPSF-LAR, and
  GDM using SVM classifiers (on
  the~\emph{lung-cancer},~\emph{breast-cancer},~\emph{leukemia},~\emph{ohsumed},~\emph{apcj-etiology},~\emph{thrombin},~\emph{kdd10},
  and~\emph{webspam} data sets) as shown in Figure~\ref{fig11}. From
  Figure~\ref{fig11}, SAOLA is very competitive with FCBF. Although
  GDM and SPSF-LAR achieve a higher prediction accuracy and a lower
  standard deviation than SAOLA, our SAOLA is highly comparable
  with those two batch methods.

\begin{figure}
\centering
\includegraphics[height=1.4in]{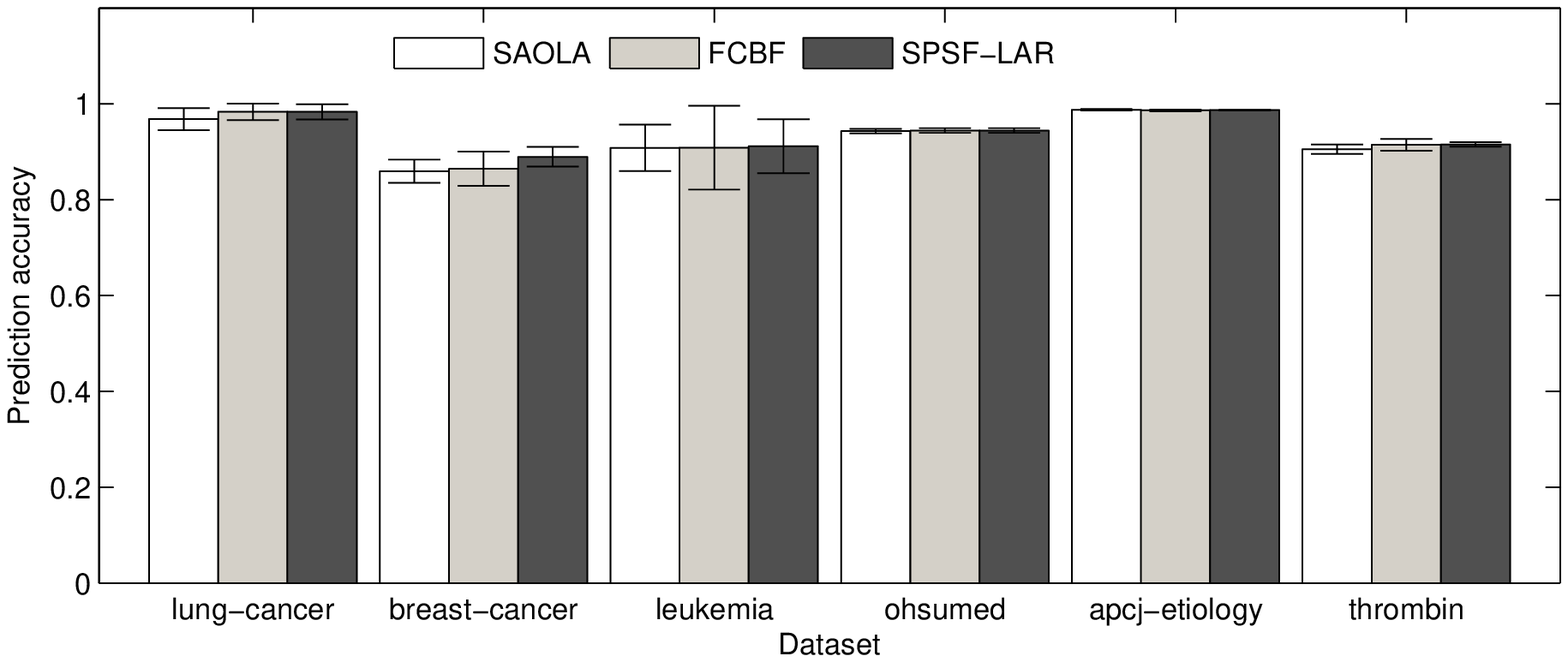}
\includegraphics[height=1.4in]{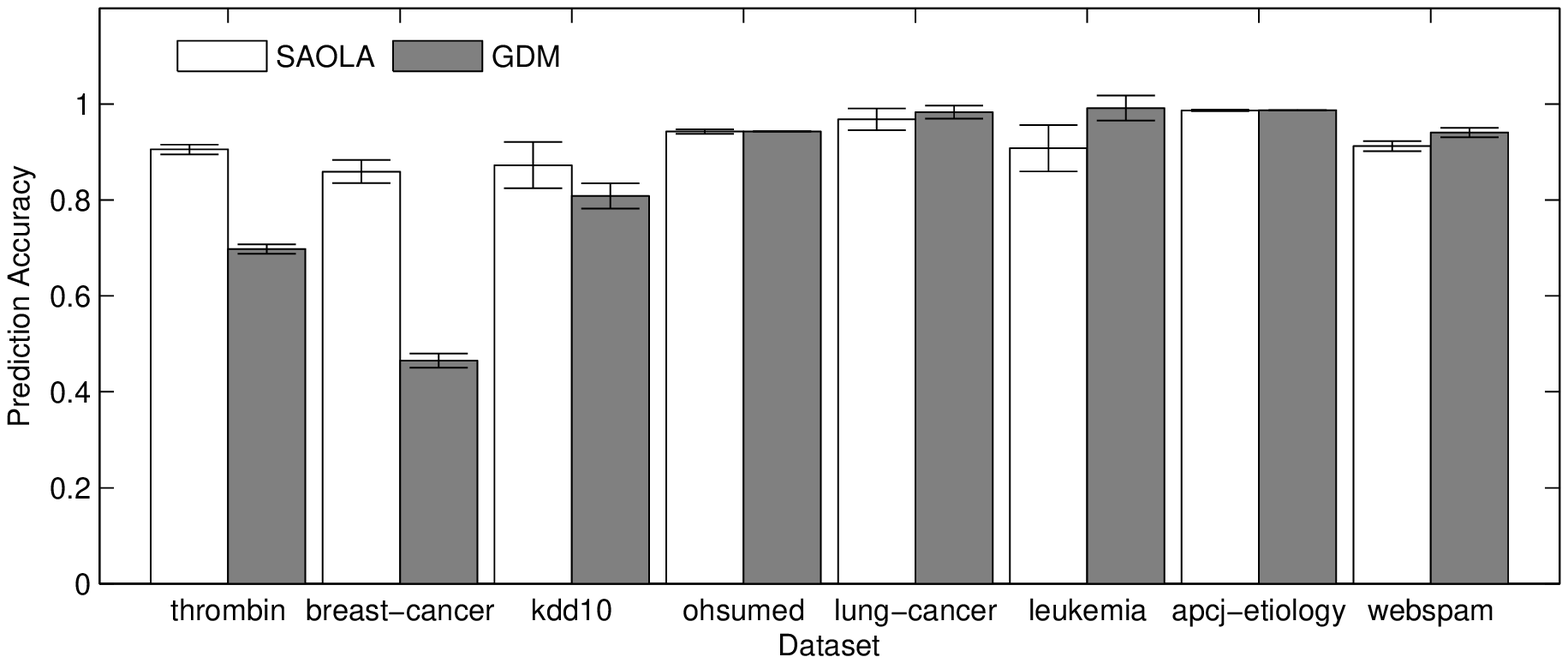}
\caption{The error bars of SAOLA, FCBF, SPSF-LAR, and GDM}
\label{fig11}
\end{figure}

\subsection{Comparison with Markov Blanket Discovery Algorithms}

In this section, we compare SAOLA and FCBF (discovery of approximate Markov blankets) with two state-of-the-art exact Markov blanket discovery algorithms, IAMB (Incremental Association Markov Blanket)~\cite{tsamardinos2003towards} and MMMB (Max-Min Markov Blanket)~\cite{tsamardinos2006max}. The IAMB algorithm finds  Markov blankets conditioned on the selected feature set currently, while the MMMB algorithm discovers Markov blankets conditioned on all possible feature subsets of the selected feature set currently\footnote{We implement IAMB and MMMB using the package of Causal Explorer at http://www.dsl-lab.org/causal\_explorer/.}. Under certain assumptions (sufficient number of data instances and reliably statistical tests), IAMB and MMMB can return the Markov blanket of a given target feature~\cite{pena2007towards,aliferis2010local}. Using the four NIPS2003 feature selection challenge data sets,~\textit{arcene} (100 data instances, 10,000 features),~\textit{dexter} (300 data instances, 20,000 features),~\textit{dorothea} (800 data instances, 100,000 features), and~\textit{gisette} (6,000 data instance, 5,000 features), we empirically study SAOLA, FCBF, IAMB, and MMMB using SVM, KNN, and J48.

\begin{table}[t]
\center
\tbl{Predicition and AUC of SAOLA, FCBF, IAMB, and MMMB (SVM) }{
\begin{tabular}{|l|l|l|l|l|l|l|l|l|}
\hline
\multirow{2}{*}{Dataset} & \multicolumn{4}{c|}{Prediction Accuracy} & \multicolumn{4}{c|}{AUC}\\ \cline{2-9} & SAOLA &FCBF  &IAMB &MMMB & SAOLA &FCBF  &IAMB &MMMB\\ \hline
arcene	&0.6600 &0.5600	&0.6000	&\textbf{0.6700} &0.6526 &0.500	&0.5917	&\textbf{0.6664}\\ \hline
dorothea	 &0.9286 &0.9171	&\textbf{0.9343} 	&-    &0.6455 &0.5735	&\textbf{0.7799}	&-\\ \hline
dexter	&\textbf{0.8500} &0.5400	&0.7600	&0.8467  &\textbf{0.8500} &0.5400	&0.7600	&0.8467\\ \hline
gisette 	&0.8950 &0.554	&0.9340	&\textbf{0.9820} &0.8950 &0.5540	&0.9340	&\textbf{0.9820}\\ \hline
\end{tabular}}
\label{new_tb1}
\end{table}

\begin{table}[t]
\center
\tbl{Predicition and AUC of SAOLA, FCBF, IAMB, and MMMB (KNN)}{
\begin{tabular}{|l|l|l|l|l|l|l|l|l|}
\hline
\multirow{2}{*}{Dataset} & \multicolumn{4}{c|}{Prediction Accuracy} & \multicolumn{4}{c|}{AUC}\\ \cline{2-9} & SAOLA &FCBF  &IAMB &MMMB & SAOLA &FCBF  &IAMB &MMMB\\ \hline
arcene	&\textbf{0.6900}	&0.5900	&0.5800	&0.6200 &\textbf{0.6867}	&0.5804	&0.5714	&0.6096\\ \hline
dorothea &\textbf{0.9200}	 &\textbf{0.9200}	&0.9143	&- &\textbf{0.7063}	&0.6932	&0.6113	&-\\ \hline
dexter	&0.7600	&0.7967	&0.6600	&\textbf{0.8233}	&0.7600	&0.7967	&0.6600	&\textbf{0.8233}	\\ \hline	
gisette	&0.8600	&0.8920	&0.9300	&\textbf{0.9690} &0.8600	&0.8920	&0.9300	&\textbf{0.9690}\\ \hline	
\end{tabular}}
\label{new_tb2}
\end{table}

\begin{table}[t]
\center
\tbl{Predicition and AUC of SAOLA, FCBF, IAMB, and MMMB (J48)}{
\begin{tabular}{|l|l|l|l|l|l|l|l|l|}
\hline
\multirow{2}{*}{Dataset} & \multicolumn{4}{c|}{Prediction Accuracy} & \multicolumn{4}{c|}{AUC}\\ \cline{2-9} & SAOLA &FCBF  &IAMB &MMMB & SAOLA &FCBF  &IAMB &MMMB\\ \hline
arcene	 &0.5600	&0.5500	&0.5800	&\textbf{0.6000} &0.5463 &0.5252	 &0.5714	&\textbf{0.5966}\\ \hline
dorothea	 &0.9343	&0.9314	&\textbf{0.9371}	&- &0.7536 &0.7258	&\textbf{0.7683}	&-\\ \hline
dexter	 &0.8133	&0.8567	&0.7833	&\textbf{0.8767} &0.8133 &0.8567	&0.7833	&\textbf{0.8767}\\ \hline
gisette	 &0.8960	&0.9130	&0.9260	&\textbf{0.9420} &0.8960 &0.9130	&0.9260	&\textbf{0.9420}\\ \hline
\end{tabular}}
\label{new_tb3}
\end{table}

From Table~\ref{new_tb1} to Table~\ref{new_tb3}, we can see that using small sample-to-feature ratio data sets, such as~\textit{arcene}, ~\textit{dexter}, ~\textit{dorothea}, SAOLA and FCBF are competitive with IAMB and MMMB, and even better than IAMB and MMMB sometimes. The explanation is that the number of data instances required by IAMB to identify a Markov blanket is at least exponential in the size of the Markov blanket since IAMB considers a conditional independence test to be reliable when the number of data instances in $D$ is at least five times the number of degrees of freedom in the test. MMMB mitigates this problem to some extent. But the performance of IAMB and MMMB may be inferior to SAOLA and FCBF as the sample-to-feature ratio becomes very small. Furthermore, as the size of a Markov blanket becomes large, MMMB is very slow due to expensive computation costs, such as on the \textit{dorothea} data set (the running time exceeds three days). But on the \textit{gisette} data set, due to the large number of data instances, IAMB and MMMB are significantly better than SAOLA and FCBF, but MMMB and IAMB take much more running time, especially MMMB.
Those empirical results conclude that the performance of SAOLA is very close to that of IAMB and MMMB on small sample-to-feature ratio data sets, but SAOLA is much more scalable than IAMB and MMMB on data sets with both many data instances and extremely high dimensionality.

\begin{table}[h!]
\center
\tbl{Number of selected features and running time of SAOLA, FCBF, IAMB, and MMMB }{
\begin{tabular}{|l|l|l|l|l|l|l|l|l|}
\hline
\multirow{2}{*}{Dataset} & \multicolumn{4}{c|}{Number of selected features} & \multicolumn{4}{c|}{Running time}\\ \cline{2-9} & SAOLA &FCBF  &IAMB &MMMB & SAOLA &FCBF  &IAMB &MMMB\\ \hline
arcene	&22	&31	&3	 &5 &3 &3	&5	 &12\\ \hline
dorothea	&63	&96	&6	 &- &58	&78	&479	&-\\ \hline
dexter	&21	&55	&4 	&11 &3	&5	 &38	&74\\ \hline
gisette	&22	&37	&9	 &308 &10	&8	 &131	 &13,483\\ \hline
\end{tabular}}
\label{new_tb4}
\end{table}

\subsection{Analysis of the Effect of Parameters on SAOLA}

\subsubsection{Analysis of Correlation Bounds}

In Section 3.3, we set the derived the correlation bound of $I(F_i;Y)$ to $min(I(F_i;C),I(Y;C))$. In this section, we investigate SAOLA using the opposite bound, i.e., $max(I(F_i;C),I(Y;C))$, which we term the SAOLA-max algorithm. In the experiments, SAOLA-max uses the same parameters as SAOLA.

Table~\ref{tb13} shows the prediction accuracies of SAOLA and SAOLA-max. With the summary of the w/t/l counts in Table~\ref{tb13}, we can see that SAOLA is very competitive with SAOLA-max in prediction accuracy. With the Friedman test at 95\% significance level,  As for SVM, SAOLA gets the higher average rank than SAOLA-max. For KNN, the null-hypothesis cannot be rejected. The average ranks calculated from the Friedman test for SAOLA and SAOLA-max are 1.46 and 1.54, respectively. With respect to J48, the average ranks for SAOLA and SAOLA-max are 1.43 and 1.57, respectively. The Friedman test testifies that SAOLA and SAOLA-max have no significant difference in prediction accuracy, although SAOLA-max gets the higher average ranks using the J48 and KNN classifiers.

\begin{table}[ht]
\centering
\tbl{Prediction accuracy}{
\begin{tabular}{|l|l|l|l|l|l|l|}
\hline
\multirow{3}{*}{Dataset} & \multicolumn{2}{c|}{KNN} & \multicolumn{2}{c|}{J48} & \multicolumn{2}{c|}{SVM}\\ \cline{2-7} & SAOLA & SAOLA-max & SAOLA &SAOLA-max & SAOLA &SAOLA-max\\ \hline
dexter                   &0.7600                &0.8000                        &0.8133                &\textbf{0.8300} &0.8500 &\textbf{0.8800}\\
lung-cancer              &\textbf{0.9833}       &0.9500           &\textbf{0.9500}          &\textbf{0.9500} &0.9833 &\textbf{1.0000}\\
hiva                     &\textbf{0.9635}       &0.9505                 &\textbf{0.9661 }      &0.9557 &\textbf{0.9635} &\textbf{0.9635}\\
breast-cancer            &0.8646                &\textbf{0.8854}         &0.7917                &\textbf{0.8125} &\textbf{0.8750}	 &0.8645\\
leukemia                 &0.9167                &\textbf{1.0000}          &\textbf{0.9583}       &\textbf{0.9583} &\textbf{0.9583}	&0.8750\\
madelon            &\textbf{0.5617 }   &\textbf{0.5617}  &\textbf{0.6083}     &\textbf{0.6083} &\textbf{0.6217	} &\textbf{0.6217}\\
ohsumed      &\textbf{0.9275}       &0.9256            &\textbf{0.9437}       &\textbf{0.9437} &\textbf{0.9431} &\textbf{0.9431}\\
apcj-etiology        &0.9793      &\textbf{0.9807}          &\textbf{0.9872}       &0.9870 &\textbf{0.9872}	&\textbf{0.9872}\\
dorothea                 &\textbf{0.9200}       &0.9171                 &\textbf{0.9343}       &0.9257  &\textbf{0.9286}	 &0.9029\\
thrombin                 &0.9374                &\textbf{0.9484}            &\textbf{0.9613}       &0.9503 &0.9116	&\textbf{0.9153}\\
news20                   &\textbf{0.7755}       &0.7592                 &0.8276                &\textbf{0.8295} &\textbf{0.8721}	&0.4993\\
url1                     &0.9627                &\textbf{0.9732}               &0.9744                &\textbf{0.9761} &\textbf{0.9645}	&0.9614\\
kdd2010          &\textbf{0.8780}    &0.8766          &0.8723           &\textbf{0.8751} &\textbf{0.8727}	&\textbf{0.8727}\\
webspam                  &0.9532                &\textbf{0.9546}          &0.9611                &\textbf{0.9635} &\textbf{0.9123}	 &0.8798\\
\hline
Ave rank                 &1.46                &\textbf{1.54}               &1.43          &\textbf{1.57} &\textbf{1.61}  &1.39\\
w/t/l                    &-                        &4/5/5                         &-                &2/10/2 &-  &5/7/2\\
\hline
\end{tabular}}
\label{tb13}
\end{table}

\begin{figure}
\centering
\includegraphics[height=1.6in,width=3.4in]{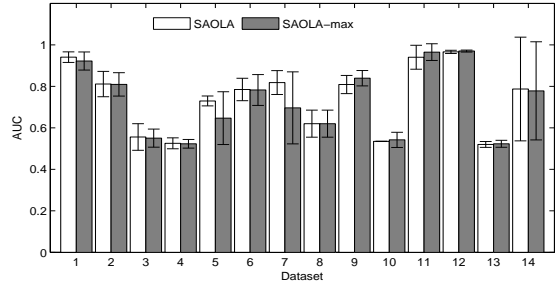}
\caption{The AUC of SAOLA and SAOLA-max  (The labels of the x-axis from 1 to 14 denote the data sets: 1. lung-cancer; 2. breast-cancer; 3. ohsumed; 4. apcj-etiology; 5. dorothea; 6. thrombin; 7. news20; 8. madelon; 9. dexter; 10. hiva; 11. leukemia; 12. url1; 13. kdd10; 14. webspam)}
\label{fig12}
\end{figure}

Moreover, Figure~\ref{fig12} shows the average AUC of
  J48, KNN and SVM and its standard deviation for SAOLA and
  SAOLA-max. We can see that SAOLA and SAOLA-max are very competitive with
  each other on all 14 data sets.

However, on the running time, Table~\ref{tb14} shows that SAOLA is much more efficient than SAOLA-max on all data sets, especially on those of extremely high dimensionality. In Table~\ref{tb14}, we can also see that SAOLA selects fewer features than SAOLA-max. The explanation is that SAOLA-max uses a bigger relevance threshold ($\delta_2=max(I(X;C),I(Y;C)$) for removing redundant features than SAOLA ($\delta_2=min(I(X;C),I(Y;C)$). Clearly, the larger the relevance threshold $\delta_2$, more features are added to the current feature set (see Steps 11 and 16 of Algorithm 1).

Compared to SAOLA-max, we can conclude that it is accurate and scalable to use the correlation bound, $\delta_2=min(I(X;C),I(Y;C)$ in the SAOLA algorithm, for pairwise comparisons to filter out redundant features.
\begin{table}[ht]
\centering
\tbl{Running time and Number of selected features}{
\begin{tabular}{|l|r|r|r|r|}
\hline
\multirow{2}{*}{Dataset} & \multicolumn{2}{l|}{Running time (seconds)}                                                                      & \multicolumn{2}{l|}{Number of selected features} \\ \cline{2-5} &SAOLA &SAOLA-max &SAOLA &SAOLA-max \\ \hline
dexter                   &\textbf{3}                        &3                         &\textbf{21}                 &39\\
lung-cancer              &\textbf{6}                        &62                        &\textbf{35}                 &260\\
hiva                     &\textbf{1}                        &3                        &\textbf{12}                 &58\\
breast-cancer            &\textbf{5}                        &40                        &\textbf{46}                 &93\\
leukemia                 &\textbf{2}                        &4                         &\textbf{17}                 &70\\
madelon                  &\textbf{0.1}           &\textbf{0.1}                      &\textbf{3}                 &\textbf{3}\\
ohsumed                  &\textbf{6}                        &8                         &\textbf{65}                 &89\\
apcj-etiology             &\textbf{22}                       &38                        &\textbf{75}                 &105\\
dorothea                 &\textbf{58}                       &327                       &\textbf{63}                 &516\\
thrombin                 &\textbf{63}                       &497                       &\textbf{20}                 &498\\
news20                   &\textbf{944}                      &2,100                      &\textbf{212}                &449\\
url1                     &\textbf{200}                      &526                       &\textbf{64}                 &346\\
kdd2010                  &\textbf{1,056}                     &2,651                      &\textbf{180}                &193\\
webspam                  &\textbf{1,456}                     &11,606                     &\textbf{51}                 &165\\
\hline
\end{tabular}}
\label{tb14}
\end{table}

Finally, we evaluate the stabilities of SAOLA and SAOLA-max using the stability measure proposed by Yu et al.~\cite{yu2008stable}. Each data set was randomly partitioned into five folds, each fold containing 1/5 of all the samples. SAOLA and SAOLA-max were repeatedly applied to four out of the five folds. This process was repeated 30 times to generate different subsamples for each data set. Then the average stabilities over 30 subsamples on each data set are as the results of SAOLA and SAOLA-max. Figure~\ref{fig13} shows the stabilities of SAOLA and SAOLA-max. We can conclude that SAOLA is very competitive with SAOLA-max on the measure of stability, although SAOLA-max can select large subsets of features.

\begin{figure}
\centering
\includegraphics[height=2in]{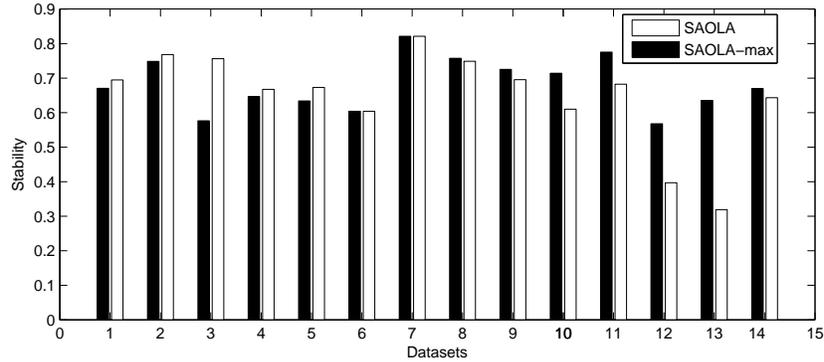}
\caption{The stabilities of SAOLA and SAOLA-max (The labels of the x-axis from 1 to 14 denote the data sets: 1. dexter; 2. lung-cancer; 3. ohsumed; 4. apcj-etiology; 5. news20; 6. breast-cancer; 7. madelon; 8. leukemia ; 9. kdd10; 10. webspam; 11. hiva; 12. dorothea; 13. thrombin; 14. url)}
\label{fig13}
\end{figure}

\subsubsection{The Effect of Input Order of Features}

Since the dimensions are presented in a sequential scan, does the input order of the features have an impact on the quality of the selected feature set? To evaluate the effect on the SAOLA algorithm, we generate 30 trials in which each trial represents a random ordering of the features in the input feature set. We apply the SAOLA algorithm to each randomized trial and report the average prediction accuracies and standard deviations (accuracy$\pm$deviation) in Table~\ref{tb_third_rev} (in the following sections, we only give the prediction accuracy of SAOLA using KNN and J48, since the prediction accuracy of SAOLA using SVM is very similar to that of SAOLA using KNN.).

\begin{table}[ht]
\centering
\tbl{Average prediction accuracy  and standard  deviation of  SAOLA}{
\begin{tabular}{|l|c|c|}
\hline
Dataset & KNN (accuracy$\pm$deviation) &J48 (accuracy$\pm$deviation) \\ \hline
ohsumed	&0.9389$\pm$0.0039	&0.9444$\pm$0.0002\\
apcj-etiology	&0.9824$\pm$0.0008	 &0.9871$\pm$0.0002\\
dorothea	&0.9193$\pm$0.0095	 &0.9345$\pm$0.0044\\
thrombin	&0.9498$\pm$0.0072	 &0.9501$\pm$0.0021\\
kdd10	&0.8759$\pm$0.0025	 &0.8682$\pm$0.0066\\
news20	&0.7694$\pm$0.0052	 &0.8194$\pm$0.0041\\
url1	&0.9557$\pm$0.0107	 &0.9729$\pm$0.0023\\
webspam	&0.9502$\pm$0.0035	 &0.9618$\pm$0.0035\\
\hline
\end{tabular}}
\label{tb_third_rev}
\end{table}

On the last eight very high-dimensional data sets, the results in Table~\ref{tb_third_rev} confirm that varying the order of the incoming features does not affect much the final outcomes. Our explanation is that with various feature orders, Steps 11 and 16 of Algorithm 1 can select the feature with the highest correlation with the class attribute among a set of correlated features and remove the corresponding correlated features of this feature.

The only difference is that in some feature orders, the final feature subset may include some weakly relevant features. For example, assuming at time $t$, $F_i$ arrives and has only one feature $Y$ that satisfies Eq.(18) in the input features, and $Y$ arrived before $F_i$ and has stayed in the currently selected feature set $S_{t_{i-1}}^*$. Then $F_i$ can be removed at time $t$ given $Y$. But if $F_i$ arrives before $Y$, and $Y$ is removed before $F_i$'s arrival, $F_i$ cannot be removed later and may be kept in the final feature set. This also explains why there is a little fluctuation of standard deviations in Table~\ref{tb_third_rev}.

\subsubsection{The Effect of $\delta$ and $\alpha$}

\begin{table}[ht]
\centering
\tbl{Prediction Accuracy}{
\tabcolsep=0.15in
\begin{tabular}{|l|c|c|c|c|}
\hline
\multirow{2}{*}{Dataset} & \multicolumn{2}{c|}{KNN} & \multicolumn{2}{c|}{J48} \\ \cline{2-5} &$\alpha$=0.01 &$\alpha$=0.05  &$\alpha$=0.01 &$\alpha$=0.05\\ \hline
madelon    &0.5617  &0.5717   &0.6083  &0.5416\\\hline
ohsumed	&0.9275 &0.9394	& 0.9437 &0.9437\\\hline
apcj-etiology &0.9793 &0.9838  &0.9872  &0.9873\\\hline
news20	&0.7755  &0.7749	 &0.8276  &0.8276\\\hline
url1	&0.9627  &0.9642	 &0.9744 &0.9744\\\hline
kdd2010	&0.8780  &0.8678	  &0.8723  &0.8723\\\hline
webspam	&0.9532  &0.9493	 &0.9611  &0.9611\\
\hline
\end{tabular}}
\label{tb15}
\end{table}

\begin{figure}[t]
\centering
\includegraphics[height=1.5in,width=3.1in]{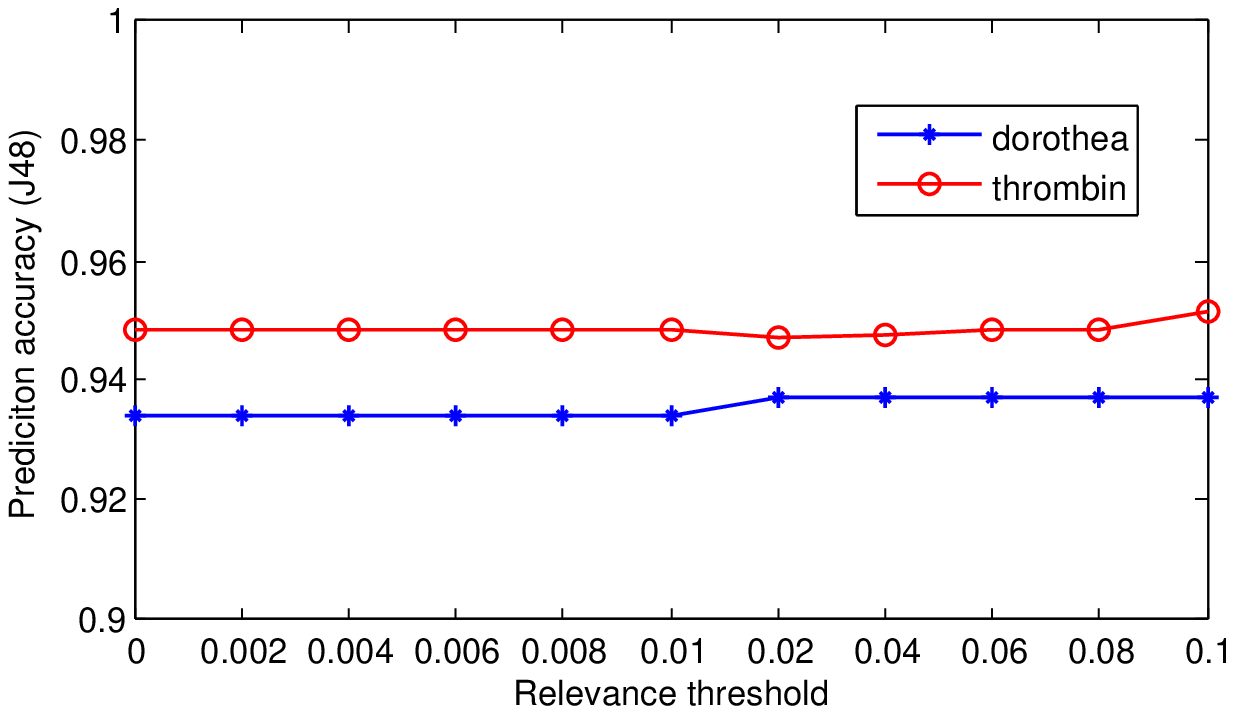}
\includegraphics[height=1.5in,width=3.1in]{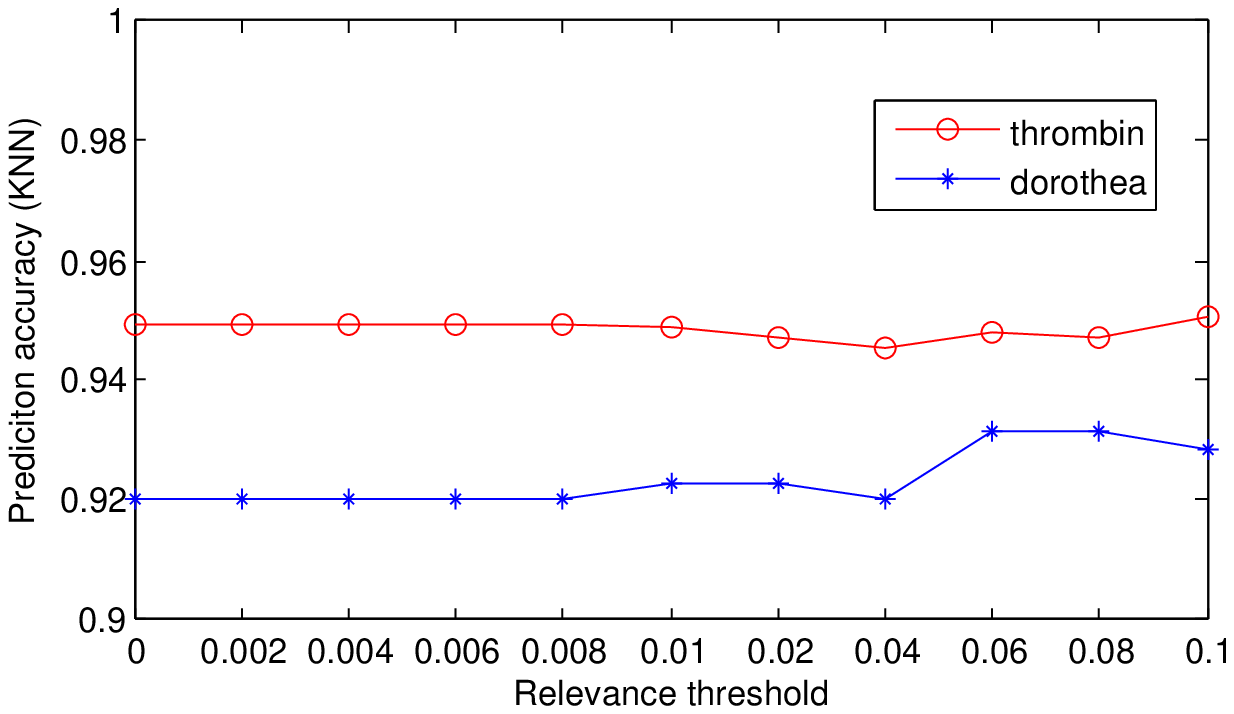}
\caption{Prediction accuracies on varied relevance thresholds (the top figure with J48 while the bottom figure with KNN)}
\label{fig16}
\end{figure}

The SAOLA algorithm has two versions: SAOLA with information gain for discrete data and SAOLA with Fisher's Z-test for continuous data. For both versions, SAOLA needs to set a relevance threshold ($\delta$ in Algorithm 1) in advance to determine whether two features are relevant. For discrete data, we set 11 different relevance thresholds for SAOLA and tuned $\delta$ using cross validation on the \emph{dorothea} and~\emph{thrombin} data sets. From Figure~\ref{fig16}, we can see that in the term of prediction accuracy, the relevance thresholds do not have a significant impact on the SAOLA algorithm.

For Fisher's Z-test, the relevance threshold is the significance level, $\alpha$, and is always set to 0.01 or 0.05. Table~\ref{tb15} shows the results of SAOLA under the different significance levels. It is clear that a significant level does not impose a significant impact on the SAOLA algorithm either.

\subsection{Comparison of Group-SAOLA with OGFS and Sparse Group Lasso}

In this section, we compare our group-SAOLA algorithm with the state-of-the-art online group feature selection methods, OGFS~\cite{wang2013online} and a well-established batch group feature selection method, Sparse Group Lasso~\cite{friedman2010note}. As for the first ten high-dimensional data sets in Table~\ref{tb2}, from \textit{dexter} to \textit{thrombin}, we randomly separated each data set into 100 feature groups without overlapping, while for the last four data sets with extremely high dimensionality, each data set was randomly separated into 10,000 feature groups without overlapping. For parameter settings of the Sparse Group Lasso algorithm, $\lambda_1\in [0.01, 0.1]$ and $\lambda_2\in [0.01, 0.1]$. For the group-SAOLA algorithm, the parameters, $\delta$ for discrete data and $\alpha$ for continuous data, are the same as the SAOLA algorithm. In the following comparisons, since the prediction accuracy of SAOLA using SVM has no significant difference than that of SAOLA using KNN and J48, we only report the prediction accuracy of SAOLA using KNN and J48.

In this section we compare group-SAOLA with OGFS and Sparse Group Lasso in terms of prediction accuracy, sizes of selected feature subsets, number of selected groups, and running time on the 14 high-dimensional data sets in Table~\ref{tb2}. We repeated this process 10 times to generate different sets of feature groups of each data set. The results as shown in Table~\ref{tb16}, Table~\ref{tb17}, Figures~\ref{fig17} to~\ref{fig20} are the average ones on those 10 sets of feature groups.

Table~\ref{tb16} summarizes the prediction accuracies
  of Group-SAOLA against OGFS and Sparse Group Lasso using the KNN and
  J48 classifiers. The highest prediction accuracy is highlighted in
  bold face. Table~\ref{tb17} illustrates the running time, sizes of
  selected feature subsets, and numbers of selected groups of
  Group-SAOLA against OGFS and Sparse Group Lasso. In
  Tables~\ref{tb16} and ~\ref{tb17}, SGLasso is the abbreviation for
  Sparse Group Lasso.

\begin{table}[ht]
\centering
\tbl{Prediction Accuracy}{
\begin{tabular}{|l|c|c|c|c|c|c|}
\hline
\multirow{2}{*}{Dataset} & \multicolumn{3}{c|}{J48} & \multicolumn{3}{c|}{KNN} \\ \cline{2-7} &group-SAOLA &OGFS  &SGLasso &group-SAOLA &OGFS  &SGLasso\\ \hline
dexter                 &0.8427   &0.5556       &\textbf{0.8800}                        &\textbf{0.7947} &0.5487  &0.7067 \\
lung-cancer          &0.9500  &0.9017      &\textbf{0.9833}                          &\textbf{0.9584}   &0.9167  &0.9333\\
hiva                      &\textbf{0.9661} &0.9602       &0.9609                         &\textbf{0.9630} &0.9471 &0.9479\\
breast-cancer         &0.6656 &0.6000    &\textbf{0.6667}                            &0.6531 &0.6385 &\textbf{0.6667}\\
leukemia               &\textbf{0.9583}  &0.7292   &\textbf{0.9583}                   &0.9833 &0.7792  &\textbf{1.0000}\\
madelon                &0.6117 &0.5153   &\textbf{0.6533}                              &0.5317 &0.4922  &\textbf{0.5967}\\
ohsumed               &\textbf{0.9439} &0.9430      &0.9431                           &0.9052   &0.9142   &\textbf{0.9431}\\
apcj-etiology          &\textbf{0.9872} &\textbf{0.9872} &\textbf{0.9872}       &0.9788    &0.9790 &\textbf{0.9818}\\
dorothea                &\textbf{0.9365} &0.9029      &0.9314                          &\textbf{0.9183} &0.8691  &0.9143 \\
thrombin                &\textbf{0.9591} &0.9396         &0.9558                       &0.9376 &0.9420  &\textbf{0.9632}\\
news20                   &\textbf{0.8188} &0.5303         &-                               &\textbf{0.7501} &0.5058   &-  \\
url1                        &\textbf{0.9715} &0.6333          &-                                 &\textbf{0.9553} &0.6089  &-  \\
kdd10                     &0.8714  &\textbf{0.8787}        &-                                   &\textbf{0.8764} &0.8758 &- \\
webspam                &\textbf{0.9341}  &0.7620         &-                                  &\textbf{0.9376} &\textbf{0.9376}  &-\\
\hline
w/t/l                      & -   &10/4/0         &0/7/3                                                         & -   &9/5/0       &3/2/5  \\
\hline
\end{tabular}}
\label{tb16}
\end{table}

\begin{table}[ht]
\centering
\tbl{Running time and Number of selected features/groups}{
\begin{tabular}{|l|r|r|r|r|r|r|}
\hline
\multirow{2}{*}{Dataset} & \multicolumn{3}{c|}{Running time (seconds)}                                                                      & \multicolumn{3}{c|}{Number of selected features/groups} \\ \cline{2-7} &group-SAOLA &OGFS &SGLasso  &group-SAOLA &OGFS &SGLasso \\ \hline
dexter                      &\textbf{2} &4  &69                                        &21/19  &72/49 &56/40\\
lung-cancer              &9 &\textbf{2}  &1,752                                    &23/19  &91/61 &384/41\\
hiva                         &2  &\textbf{1} &734                                        &5/5  &47/38 &55/19\\
breast-cancer           &8 &\textbf{3}   &275                                      &8/7 &111/61 &2,775/30\\
leukemia                  &3 &\textbf{1}   &18                                       &16/14 &66/47 &61/28\\
madelon                  &\textbf{0.1} &\textbf{0.1} &27                        &2/2 &15/13 &5/2\\
ohsumed                 &\textbf{2} &6       &64                                    &17/16 &61/43 &385/28\\
apcj-etiology            &\textbf{34} &\textbf{34} &210                        &47/33 &44/44 &44/12\\
dorothea                 &23 &\textbf{22}     &112                                  &41/27 &126/67 &135/6\\
thrombin                 &\textbf{39} &1,015  &1,015                             &7/5 &691/84 &691/70\\
news20                   &2,154 &\textbf{1,054} &-                              &140/140 &192/192 &-\\
url                          &\textbf{306} &3,598   &-                                &29/29 &73/73 &-\\
kdd10                     &\textbf{395} &39,213   &-                              &52/52 &133/132 &-\\
webspam                &\textbf{3,013}  &20,718  &-                            &17/17  &401/395 &-\\
\hline
\end{tabular}}
\label{tb17}
\end{table}

\subsubsection{Comparison of Group-SAOLA with OGFS}

In this section we compare OGFS with group-SAOLA on the 14 high-dimensional data sets in Table~\ref{tb2}, and the results are as shown in Tables~\ref{tb16} to~\ref{tb17}, Figures~\ref{fig17} to~\ref{fig20}. In Tables~\ref{tb16}, from the the win/tie/lose counts in the last rows of the table, we observe that Group-SAOLA never loses against OGFS on all of the 14 high-dimensional data sets.

To evaluate whether the prediction accuracy of group-SAOLA and that of OGFS have no significant difference, using the Friedman test, for the J48 classifier, the null-hypothesis is rejected, and the average ranks for group-SAOLA and OGFS are 1.8929 and 1.1071, respectively. Then we proceed with the Nemenyi test as a post-hoc test, and the critical difference is up to 0.6885. The difference between 1.8929 and 1.1071 is bigger than this critical difference, then group-SAOLA is significantly better than OGFS in prediction accuracy.

For the KNN classifier, the null-hypothesis cannot be rejected. The average ranks for group-SAOLA and OGFS are 1.75 and 1.25, respectively. Accordingly, for KNN, group-SAOLA and OGFS have no significant difference in prediction accuracy.

Figure~\ref{fig17} gives the error bars (the left figure) and
   AUC (the right figure) of group-SAOLA and OGFS using the KNN
  classifier. We can see that group-SAOLA clearly outperforms OGFS
  using the AUC and error bar metrics.

\begin{figure}
\centering
\includegraphics[height=1.55in,width=2.7in]{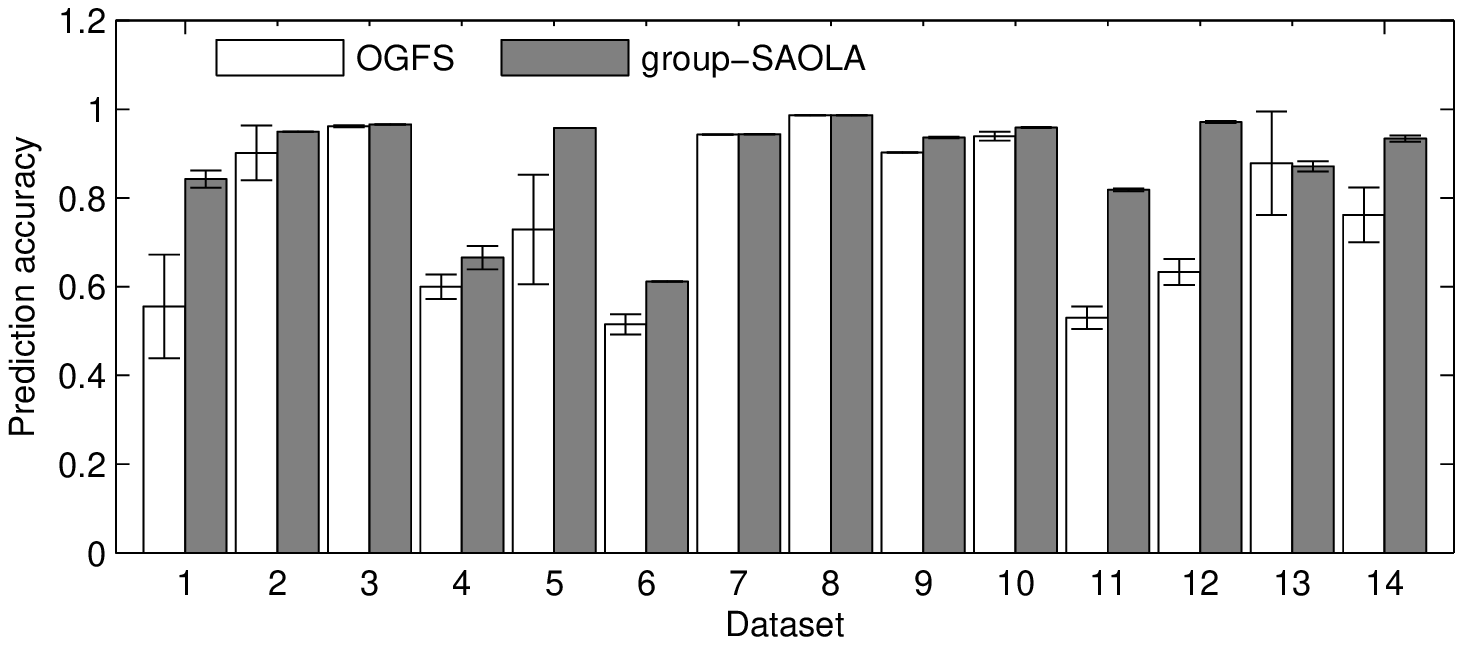}
\includegraphics[height=1.55in,width=2.7in]{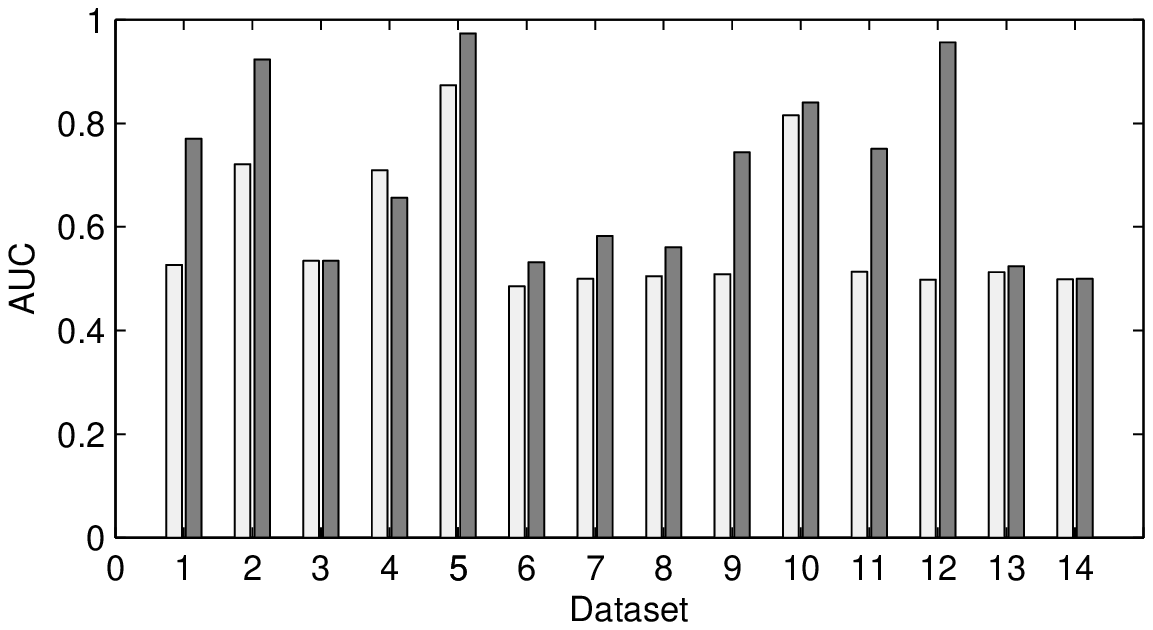}
\caption{The error bars and AUC of group-SAOLA and OGFS (The labels of the x-axis from 1 to 14 denote the data sets: 1. dexter; 2. lung-cancer; 3. hiva; 4. breast-cancer; 5. leukemia; 6. madelon; 7. ohsumed; 8. apcj-etiology; 9. dorothea; 10. thrombin; 11. news20; 12. url1; 13. kdd10; 14. webspam)}
\label{fig17}
\end{figure}

Furthermore, Table~\ref{tb17} illustrates that group-SAOLA is faster
than OGFS on most data sets. When the number of features increases to
millions and the number of feature groups becomes large, OGFS becomes
very costly, but our group-SAOLA is still scalable and efficient. The
explanation is that the time complexity of group-SAOLA is determined
by the number of features within the currently selected feature
groups, and the strategy of online redundancy detection within the
currently selected feature groups makes group-SAOLA very
scalable. Meanwhile, from Table~\ref{tb17}, we observe that
group-SAOLA selects fewer features than OGFS. From the
  selected numbers of groups and selected numbers of features, we can
  see that group-SAOLA not only selects the smaller number of feature
  groups, but also achieves more sparsity of within groups than
  OGFS.

Figure~\ref{fig18} gives the results of the numbers of selected groups and the corresponding prediction accuracy for group-SAOLA, and OGFS using the KNN classifier on the 14 data sets in Table~\ref{tb2}. The best possible mark for each graph is at the upper left corner, which selects the fewest groups with the highest prediction accuracy. We can see that group-SAOLA selects fewer groups while gets higher prediction accuracies than OGFS on all the 14 data sets in Table~\ref{tb2}, except for the~\textit{ohsumed} data set. However, on the~\textit{ohsumed} data set, on the prediction accuracy, group-SAOLA is very competitive with OGFS.

\begin{figure}
\centering
\includegraphics[height=2.5in]{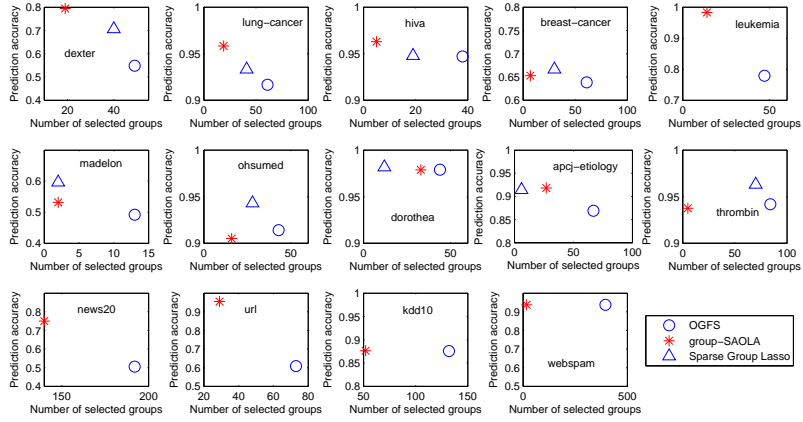}
\caption{Accuracies and numbers of selected groups of three algorithms (KNN)}
\label{fig18}
\end{figure}

Figure~\ref{fig19} gives the results of the numbers of selected groups and the corresponding prediction accuracy for Group-SAOLA and OGFS using the J48 classifier. We can see that Group-SAOLA selects fewer groups while gets higher prediction accuracies than OGFS on all the fourteen data sets.

\begin{figure}
\centering
\includegraphics[height=2.5in]{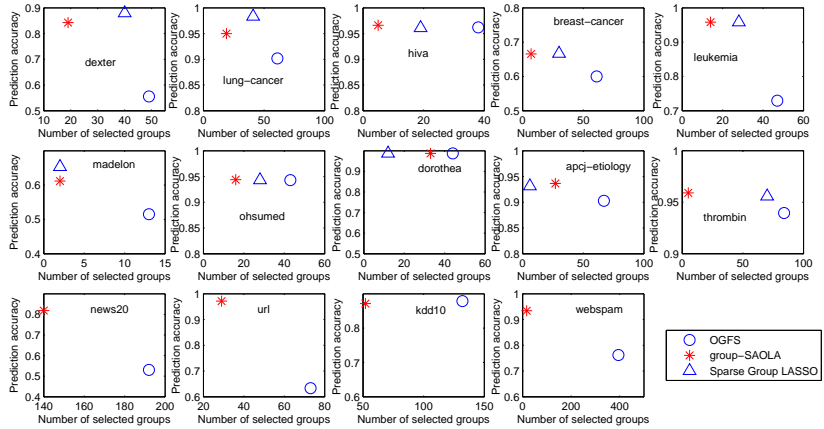}
\caption{Accuracies and numbers of selected groups of three algorithms (J48)}
\label{fig19}
\end{figure}

\subsubsection{Comparison of Group-SAOLA with Sparse Group Lasso}

Since Sparse Group Lasso\footnote{The codes are available at http:$//$yelab.net/software/SLEP/} can only deal with the first ten high-dimensional data sets in Table~\ref{tb2} due to high computational costs, in this section we compare it with group-SAOLA on those first ten high-dimensional data sets.

With Table~\ref{tb16}, using the Friedman test, for the J48 classifier, the null-hypothesis cannot be rejected, and the average ranks for group-SAOLA and Sparse Group Lasso are 1.5 and 1.5, respectively. For the KNN classifier, the null-hypothesis is accepted, and the average ranks for group-SAOLA and Sparse Group Lasso are 1.6 and 1.4, respectively. Accordingly, for the J48 and KNN classifiers, group-SAOLA and Sparse Group Lasso have no significant difference in prediction accuracy. Furthermore,
Table~\ref{tb17} shows that group-SAOLA is much faster than Sparse Group Lasso, and group-SAOLA selects fewer features than Sparse Group Lasso.

Figure~\ref{fig20} gives the AUC (the left figure) and error bars (the right figure) of group-SAOLA and Sparse Group Lasso using the KNN classifier. Figure~\ref{fig20} illustrates that group-SAOLA is very competitive with Sparse Group Lasso using the AUC and error bar metrics.

\begin{figure}
\centering
\includegraphics[height=1.55in,width=2.7in]{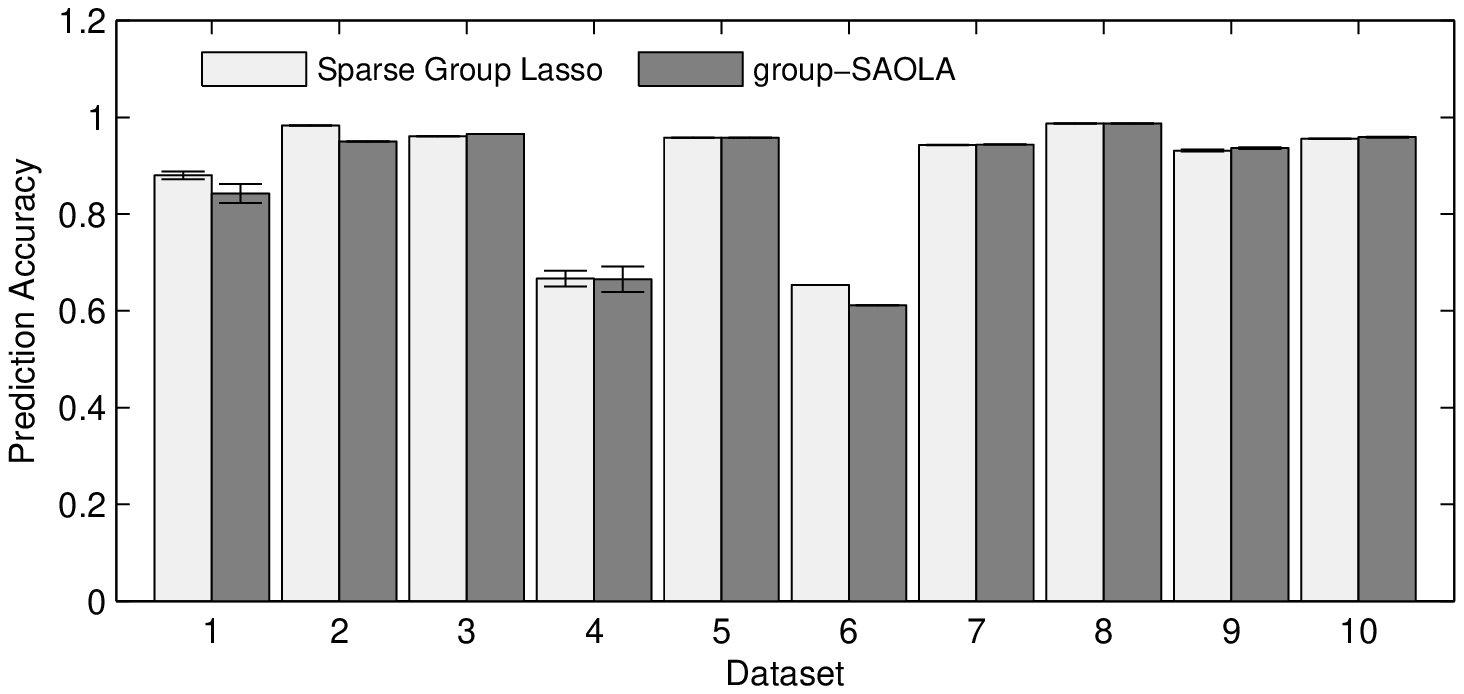}
\includegraphics[height=1.55in,width=2.7in]{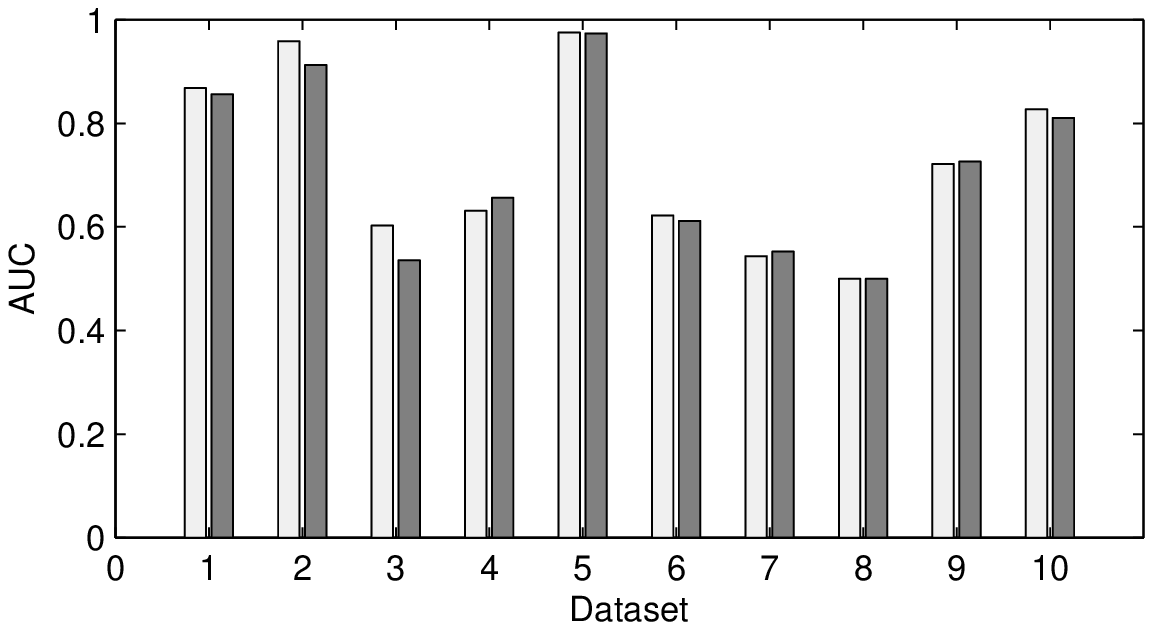}
\caption{The error bars  and AUC of group-SAOLA and Sparse Group Lasso (The labels of the x-axis from 1 to 10 denote the data sets: 1. dexter; 2. lung-cancer; 3. hiva; 4. breast-cancer; 5. leukemia; 6. madelon; 7. ohsumed; 8. apcj-etiology; 9. dorothea; 10. thrombin)}
\label{fig20}
\end{figure}

Figures~\ref{fig18} and~\ref{fig19} give the results of the numbers of selected groups and the corresponding prediction accuracy for group-SAOLA and Sparse Group Lasso using the KNN and J48 classifiers, respectively. The best possible mark for each graph is at the upper left corner. We can see that group-SAOLA prefers to select few groups, in comparison to Sparse Group Lasso. Meanwhile, group-SAOLA is very competitive with Sparse Group Lasso in terms of prediction accuracy.

In summary, the group-SAOLA algorithm is a scalable and accurate online group feature selection approach. This validates that without requiring a complete set of feature groups on a training data set before feature selection starts, group-SAOLA is very competitive comparing to the well-established the Sparse Group Lasso algorithm.

\section{Conclusions and Future Work}

In this paper, we presented the SAOLA algorithm, a scalable and
accurate online approach to tackle feature selection with extremely
high dimensionality. We conducted a theoretical analysis and derived a
lower bound of correlations between features for pairwise comparisons,
and then proposed a set of online pairwise comparisons to maintain a
parsimonious model over time. To deal with the group structure
information in features, we extended the SAOLA algorithm,
and then proposed a novel group-SAOLA algorithm to deal with features that arrive by groups. The group-SAOLA algorithm can
online maintain a set of feature groups that is sparse between groups
and within each group simultaneously.

Using a series of benchmark data sets, we compared the SAOLA and
group-SAOLA algorithms with state-of-the-art online feature
selection methods and well-established batch feature selection
algorithms. The empirical study demonstrated that the SAOLA and
group-SAOLA algorithms are both scalable on data sets of extremely
high dimensionality, have superior performance over
state-of-the-art online feature selection methods, and are very
competitive with state-of-the-art batch feature selection methods
in prediction accuracy, while much faster in running time.

In this work, we have used online pairwise comparisons
  to calculate the correlations between features without further
  exploring positive feature interactions between features. Moreover,
  from the AUC results reported in the work, we can see that SAOLA and
  its rivals, including the three online algorithms and three batch
  methods, cannot effectively deal with class-imbalanced
  data. Thus, we will further explore the
  following directions in online feature selection: efficient and
  effective methods to discover positive feature interactions between
  features, and accurate and scalable online algorithms to handle
  class-imbalanced data.

Meanwhile, our empirical studies have validated that SAOLA (using pairwise comparisons) is competitive with IAMB and MMMB (using multiple comparisons). To conduct thoroughly theoretical analysis and empirical studies on why pairwise feature correlations (instead of conditioning on all possible feature subsets) may be sufficient in practice deserve further exploration in our future work.

\section*{Acknowledgments}
{
The preliminary version of this manuscript with the title ``Towards Scalable and Accurate Online Feature Selection for Big Data'' was published in the proceedings of
14th IEEE International Conference on Data Mining (ICDM2014), 660-669. This work is partly supported by a PIMS Post-Doctoral Fellowship Award of the Pacific Institute for the
Mathematical Sciences, Canada.
}

\bibliographystyle{ACM-Reference-Format-Journals}
\bibliography{sigproc}

\end{sloppy}
\end{document}